\documentclass[10pt,twocolumn,letterpaper]{article}

\usepackage[pagenumbers]{cvpr} 

\usepackage{amsmath,amsfonts,bm}

\def\eqref#1{equation~\ref{#1}}

\def\1{\bm{1}}

\DeclareMathAlphabet{\mathsfit}{\encodingdefault}{\sfdefault}{m}{sl}
\SetMathAlphabet{\mathsfit}{bold}{\encodingdefault}{\sfdefault}{bx}{n}

\usepackage{graphicx}
\usepackage{amsmath}
\usepackage{amssymb}
\usepackage{mathtools} 

\usepackage{xcolor}
\usepackage{enumitem}
\usepackage{xspace}
\usepackage{booktabs}
\usepackage{colortbl}
\usepackage{multirow}
\usepackage{makecell}
\usepackage{lipsum} 
\usepackage{gensymb} 
\usepackage{wrapfig} 

\newcommand{\Tref}[1]{Table~\ref{#1}}

\newcommand{\fref}[1]{Fig.~\ref{#1}}
\newcommand{\Fref}[1]{Figure~\ref{#1}}
\newcommand{\sref}[1]{Sec.~\ref{#1}}

\usepackage[labelsep=period]{caption}
\renewcommand{\paragraph}[1]{\vspace{0.2em}\noindent \textbf{#1 \hspace{0.2em}}}

\definecolor{MyDarkRed}{rgb}{0.46, 0.16, 0.16}
\definecolor{MyDarkBlue}{rgb}{0.16, 0.16, 0.66}

\newcommand{\modelname}{\mbox{LHM++}\xspace} 
\newcommand{\MethodName}{LHM++\xspace}

\newcommand{\Frst}[1]{\textcolor{red}{\textbf{#1}}}
\newcommand{\Scnd}[1]{\textcolor{blue}{\textbf{#1}}}

\definecolor{cvprblue}{rgb}{0.21,0.49,0.74}
\usepackage[pagebackref=true,breaklinks=true,colorlinks=true,bookmarks=false,linkcolor={red!50!black},urlcolor={darkgray!80!black},citecolor={green!50!black}]{hyperref} 
\usepackage{url}

\usepackage{mathtools}

\usepackage{float}
\usepackage{enumitem}
\usepackage{comment}
\usepackage{pifont}

\usepackage{soul}
\usepackage{acronym}
\usepackage{fancybox}
\usepackage{epigraph}
\usepackage{dirtytalk}
\usepackage{bm}
\usepackage{fontawesome}
\usepackage{colortbl}

\def\paperID{1131} 
\def\confName{CVPR}
\def\confYear{2026}

\title{\MethodName: An Efficient Human Reconstruction Model for Pose-free Images to 3D}

\author{Lingteng Qiu$^{1}\footnotemark[1]$ \quad Peihao Li$^{1}\footnotemark[1]$  \quad Heyuan Li$^{3}\footnotemark[1]$ \\
Qi Zuo$^{1}$ \quad Xiaodong Gu$^{1}$ \quad Yuan Dong$^{1}$ \quad Weihao Yuan$^{1}$ \quad Rui Peng$^{1}$ \quad Siyu Zhu$^{5}$ \quad \\ Xiaoguang Han$^{3,4}$ Guanying Chen $^{2}\footnotemark[2]$ \quad Zilong Dong $^{1}\footnotemark[2]$ \\
{\normalsize $^1$Tongyi Lab, Alibaba Group}
\quad{\normalsize $^2$Sun Yat-sen University} \\ {\normalsize $^3$SSE, CUHKSZ} \quad {\normalsize $^4$FNii, CUHKSZ} \quad{\normalsize $^5$Fudan University} 
}

\begin{document}

\twocolumn[{
\renewcommand\twocolumn[1][]{#1}
\maketitle

\vspace{-18pt}
\begin{center}
    \captionsetup{type=figure}
    \includegraphics[width=\textwidth]{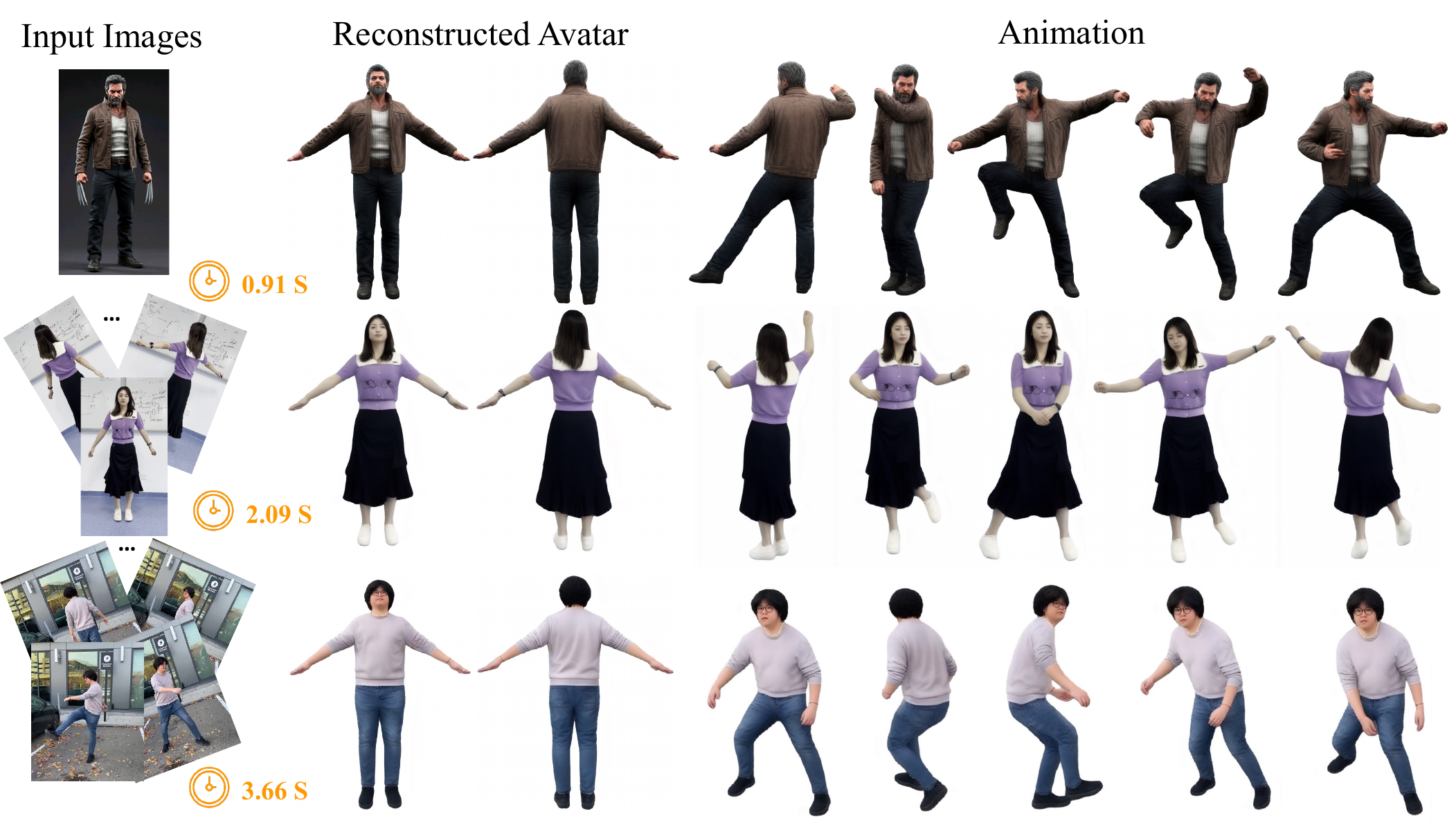}
        \captionof{figure}{\textbf{3D Avatar Reconstruction and Animation Results of our \modelname.} Given a set of $N \ge 1$ images of a human subject, without requiring camera parameters or human pose annotations,
    our method can reconstruct a high-fidelity, animatable 3D human avatar in seconds.
    }
    \label{fig:teaser}
\end{center}
}]

\definecolor{linkred}{RGB}{255, 30, 50} 

\begin{abstract}
Reconstructing animatable 3D humans from casually captured images of articulated subjects without camera or pose information is highly practical but remains challenging due to view misalignment, occlusions, and the absence of structural priors.
In this work, we present \emph{\modelname}, an efficient large-scale human reconstruction model that generates high-quality, animatable 3D avatars within seconds from one or multiple pose-free images. 
At its core is an \emph{Encoder–Decoder Point–Image Transformer} architecture that progressively encodes and decodes 3D geometric point features to improve efficiency, while fusing hierarchical 3D point features with image features through multimodal attention. 
The fused features are decoded into 3D Gaussian splats to recover detailed geometry and appearance.
To further enhance visual fidelity, we introduce a lightweight 3D-aware neural animation renderer that refines the rendering quality of reconstructed avatars in real time.
Extensive experiments show that our method produces high-fidelity, animatable 3D humans without requiring camera or pose annotations. 
Our code and project page are available at \href{https://lingtengqiu.github.io/LHM++/}{\textcolor{linkred}{https://lingtengqiu.github.io/LHM++/}}.
\end{abstract}

\section{Introduction}
\label{sec:intro}
Reconstructing high-quality, animatable 3D human avatars from casually captured images is a crucial task in computer graphics, with broad applications like virtual reality and telepresence. 
A practical solution should support rapid and robust reconstruction from minimal input—ideally using only one or a few casually captured images, without relying on camera parameters, human pose annotations, or controlled capture environments.
Such capability is essential for enabling scalable and accessible avatar generation in real-world scenarios.

Existing approaches to animatable 3D human reconstruction from monocular or multi-view videos typically rely on optimization-based frameworks that minimize photometric or silhouette reprojection losses \cite{alldieck2018videobasedreconstruction3d,weng2022humannerf,guo2025vid2avatar}. These methods usually require dozens or hundreds of images with accurate human pose estimation as a prerequisite. Moreover, the optimization process is often computationally expensive, taking several minutes or even hours to converge, thus limiting real-time applications.

More recently, LHM~\cite{qiu2025LHM}, a feed-forward network for single-image 3D human reconstruction, has shown promising progress toward real-time performance. 
It employs a transformer-based architecture to fuse geometric point features initialized from the canonical SMPL-X surfaces and image features to directly predict a 3D Gaussian Splatting~\cite{kerbl3Dgaussians} based avatar from a single image. 
However, as single-image-based methods are inherently limited by partial observations, they often struggle to reconstruct occluded or unseen regions, leading to oversmoothed surfaces or noticeable artifacts~\cite{saito2020pifuhd,zhuang2024idol}.
 
A straightforward extension of LHM~\cite{qiu2025LHM} to multi-image settings would involve concatenating image tokens from multiple images and performing attention fusion. 
However, such a naive approach suffers from substantial memory and computational overhead due to the large number of geometric point features and the quadratic complexity of dense self-attention mechanisms.

In this work, we propose \emph{\MethodName}, a novel feed-forward framework for fast and high-fidelity 3D human reconstruction from one or a few images without requiring the camera and human poses. 
To achieve this, we design an efficient \emph{Encoder-Decoder Point-Image Transformer (PIT) Framework} that hierarchically fuses 3D geometric features with multi-image cues. 
The framework is built upon \emph{Point-Image Transformer blocks (PIT-blocks)}, which enable interaction between geometric and image tokens via attention fusion while maintaining scalability through spatial hierarchy.

We start by representing the SMPL-X anchor points as geometric tokens and extracting image tokens from each input image. 
The encoder stage comprises several PIT-blocks to progressively downsample the geometric tokens via Grid Pooling~\cite{wu2022ptv2}. 
At each layer, the downsampled point tokens interact with image tokens through multimodal attention~\cite{esser2024scaling}, allowing compact yet expressive geometric representations to be enriched with visual information from multiple images.
The decoder stage upsamples the geometric tokens to recover spatial resolution. 

The resulting 3D geometry tokens are decoded to predict Gaussian splatting parameters, enabling high-quality rendering and animation.
To further improve visual fidelity, we introduce a lightweight 3D-aware neural animation renderer, built on a DPT-head architecture, that refines the rendering quality of reconstructed avatars in real time.
For robustness and generalization, we train our model on large-scale real-world human video datasets spanning diverse clothing styles, body shapes, and viewing conditions.
In summary, our contributions are:
\begin{itemize}[leftmargin=*,topsep=0pt]
    \item We introduce \emph{\MethodName}, an efficient feed-forward model, capable of reconstructing high-quality and animatable 3D human avatars in seconds from one or a few casually captured images, without requiring either camera poses or human pose annotations.
    \item We propose a novel \emph{Encoder-Decoder Point-Image Transformer} (PIT) architecture that hierarchically fuses 3D geometric point features and 2D image features using multimodal attention, enabling efficient and scalable integration of multi-image cues.
    \item Extensive experiments on both synthetic and real-world data demonstrate that \MethodName\ unifies single- and multi-image 3D human reconstruction, with superior generalization and visual quality.
\end{itemize}



\section{Related Work}
\label{sec:related_work }

\paragraph{Human Reconstruction from A Single Image}
For single-image 3D human reconstruction, many methods adopt implicit neural representations~\cite{saito2019pifu,saito2020pifuhd,xiu2023econexplicitclothedhumans,cao2022jiff,zheng2020pamirparametricmodelconditionedimplicit,zhang2024sifusideviewconditionedimplicit,xiong2024mvhumannet,yang2024have} to model complex human geometries. To improve geometric consistency and generalizability, some approaches~\cite{choutas2022accurate3dbodyshape,kanazawa2018endtoendrecoveryhumanshape,alldieck2018detailedhumanavatarsmonocular,alldieck2019tex2shapedetailedhumanbody} rely on parametric body models such as SMPL~\cite{loper2015smpl,smplx:2019} to predict geometric offsets for the reconstruction of clothed humans. However, reconstruction from a single image is an ill-posed problem. Current cascade-type approaches~\cite{li2024pshuman,wang2025wonderhuman,weng2024template,wang2025humandreamer,qiu2024AniGS} attempt to mitigate this issue by decoupling the process into two stages: multi-view image synthesis using generative models, followed by 3D reconstruction. While these methods require view-consistent generation in the first stage, which is often unstable and challenging, this ultimately affects the quality of the reconstruction.

Inspired by the success of large reconstruction models~\cite{hong2023lrm,tang2025lgm}, emerging solutions aim to enable direct generalizable reconstruction through feed-forward networks which significantly accelerate the inference time. Human-LRM~\cite{weng2024template} employs a feed-forward model to decode the triplane NeRF representation, then followed by a conditional diffusion-based novel views generation and reconstruction. IDOL~\cite{zhuang2024idol} introduces a UV-Alignment transformer model to decode Gaussian attribute maps in a structured 2D UV space. LHM~\cite{qiu2025LHM} leverages a Body-Head multimodal transformer architecture produces animatable 3D avatars with the face identity preservation and fine detail recovery.  While these single-view methods often face challenges with occlusions and invisible regions, frequently resulting in geometrically implausible results or blurred textures. 

Recently, HumanRAM~\cite{yu2025humanram} adapts human reconstruction to novel view and pose synthesis using LVSM with static pose sparse view inputs, achieving photorealistic results but suffering from slow transformer-based rendering.
The concurrent work, GIGA~\cite{zubekhin2025giga}, employs UV map representations for generalizable reconstruction, but it relies on multi-view captures of the same action, complex camera setups, and motion calibration.

\paragraph{Human Reconstruction from Monocular Videos}
Video-based techniques further improve reconstruction consistency by using temporal cues. 4D replay methods~\cite{park2021hypernerf, weng2022humannerf} can reconstruct dynamic humans from monocular video or multiview video sequences, however, they are not able to drive the humans in novel poses since they do not build a standalone 3D model for humans. Therefore, a series of monocular video-based methods~\cite{jiang2022selfrecon,qiu2023recmv,hu2023gauhuman,tan2025dressrecon} build a static 3D human model and can drive the human in novel poses by binding the skinning weight.

\begin{table}[tb] \centering
    \caption{Comparison with state-of-the-art 3D human reconstruction methods. FF stands for Feed-forward, PF for Pose-free, and AM for Animatable.}
    \label{tab:compare}
    \resizebox{0.48\textwidth}{!}{
        \begin{tabular}{l*{5}{c}}
        \toprule
        Method & \# Image & FF & PF & AM & Runtime \\    
        \midrule
        CAR~\cite{liao2023high} & $1$ & \ding{56} & \ding{56} & \ding{52} & 5 Minutes \\
        IDOL~\cite{zhuang2024idol} & $1$ & \ding{52} &\ding{52} &\ding{52} & Seconds \\
        AniGS~\cite{qiu2024AniGS} & $1$ & \ding{56} &\ding{52} &\ding{52} & 15~Minutes \\
        LHM~\cite{qiu2025LHM} & $1$ & \ding{52} &\ding{52} &\ding{52} & Seconds \\
        \midrule
        Vid2Avatar~\cite{guo2023vid2avatar} & $>100$ & \ding{56} &\ding{56} &\ding{52} & 1-2 Days \\
        Hugs~\cite{kocabas2024hugs} & $>80$ & \ding{56} & \ding{56} & \ding{52} & 30 Minutes \\
        Canonicalfusion~\cite{shin2024canonicalfusion} & $>1$ & \ding{56} & \ding{56} & \ding{52} & 11 Minutes \\
        GPS-Gaussian~\cite{zheng2024gpsgaussian} & $2$ & \ding{52} &\ding{56} &\ding{56} & Seconds \\
        3DGS-Avatar~\cite{qian20233dgsavatar} & $>20$ & \ding{56} &\ding{56} &\ding{52} & 0.5 Hours \\
        InstantAvatar~\cite{jiang2023instantavatar} & $>20$ & \ding{56} &\ding{56} &\ding{52} & Minutes \\
        GaussianAvatar~\cite{hu2024gaussianavatar} & $>20$ & \ding{56} &\ding{56} &\ding{52} & 0.5-6 Hours \\
        ExAvatar~\cite{moon2024exavatar}  & $>20$ &\ding{56} &\ding{56} &\ding{52} & 1.5-5 Hours \\
        PuzzleAvatar~\cite{xiu2024puzzleavatar} & $ 4\sim6$ & \ding{56} &\ding{52} &\ding{56} & 4-6 Hours \\
        Vid2Avatar-Pro~\cite{guo2025vid2avatar}  & $>100$ &\ding{56} &\ding{56} &\ding{52} & Hours \\
        Giga~\cite{zubekhin2025giga} & $1\sim4$ & \ding{52} & \ding{56} & \ding{52} & Seconds\\ 
        FRESA~\cite{wang2025fresa} & $1\sim4$ & \ding{52} & \ding{52} & \ding{52} & Seconds \\
        \midrule
        Ours & $\ge 1$ &\ding{52} &\ding{52} &\ding{52} & Seconds \\
        \bottomrule
    \end{tabular}
    }
\end{table}

Another series of works~\cite{jiang2023instantavatar,moon2024exavatar,hu2024gaussianavatar,guo2025vid2avatar,jiang2022neuman,qian20243dgs,yu2023monohuman,zhan2025tomie, yang2025sigman, kocabas2024hugs} take it further by incorporating a 3D parametric human model into the optimization process, and thus can drive the human reconstruction in novel poses without any post-processing. Despite impressive visual fidelity, they often require dozens of minutes and dozens of views for a good optimization, which limits their practical usage in real-world scenarios.

Unconstrained collection is ideal input for a practical application. However, existing methods~\cite{xiu2024puzzleavatar,yang2024have} share a similar pipeline that uses a view generative model and score distillation sampling~\cite{poole2022dreamfusion} for shape optimization. As a result, they are costly for offline training and impractical for online reconstruction. 

\paragraph{Feed-Forward Scene Reconstruction}
Recent years have witnessed a paradigm shift in geometric 3D vision, driven by the emergence of methods that eliminate traditional dependencies on camera calibration and multi-stage pipelines. At the forefront of this revolution lies the DUSt3R~\cite{wang2024dust3r} framework, which reimagines 3D reconstruction as a direct regression problem from image pairs to 3D pointmaps. By discarding the need for intrinsic camera parameters, extrinsic pose estimation, or even known correspondence relationships, DUSt3R and its successors~\cite{yang2025fast3r,tang2024mv,lu2024align3r,wang2025vggt} have democratized 3D vision, enabling rapid reconstruction across diverse scenarios while achieving state-of-the-art performance in depth estimation, relative pose recovery, and scene understanding. However, general feed-forward reconstruction methods assume that images are captured from a static scene~\cite{wang2023pf,li2024m}, while our \modelname can accept human images with different camera and human poses as input and produce an animatable 3D avatar. The concurrent work, FastVGGT~\cite{shen2025fastvggt}, employs a token merge strategy to improve the efficiency of image token global attention.

\begin{figure*}[tb] \centering
    \includegraphics[width=\textwidth]{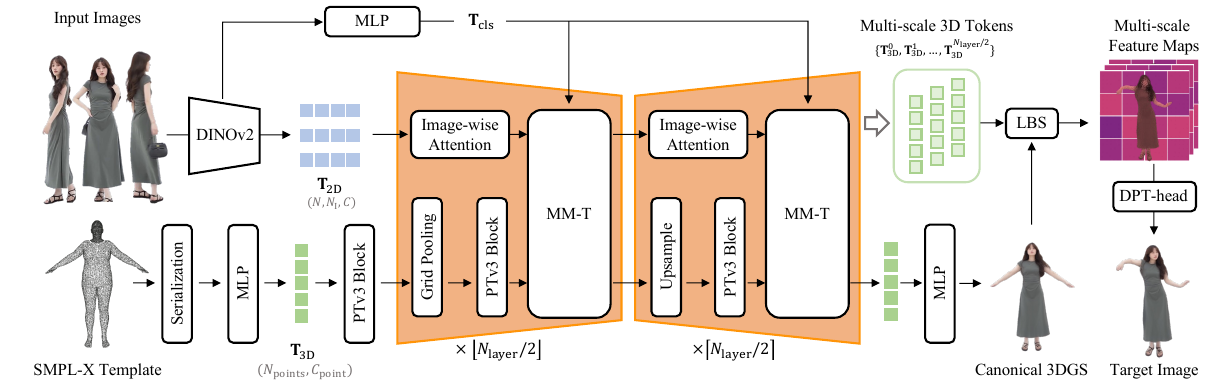}
    \caption{\textbf{Overview of the proposed \emph{\MethodName}}. In 2D space, we extract image tokens $\mathbf{T}_{\text{2D}}$ from input RGB images by DINOv2. In 3D space, geometric tokens $\mathbf{T}_{\text{3D}}$ are derived from SMPL-X anchor points via an MLP.
    Next, we design an Encoder-Decoder Point-Image Transformer (PIT) to hierarchically fuse 3D and 2D tokens, where the downsampled 3D tokens interact with 2D tokens via multi-modal attention in each layer.
    The final 3D tokens are decoded to predict 3D Gaussian parameters, followed by a light-weight DPT head for photorealistic animation. 
    } \label{fig:pipeline}
\end{figure*}

\section{Method}
\subsection{Overview}
\label{sub:Overview}

\paragraph{Problem Formulation} 
Given a set of  $N \geq 1$ RGB images $I^1,\dots,I^{N}$ of a human subject, without known camera parameters or human pose annotations, our goal is to reconstruct a high-fidelity, animatable 3D human avatar $\chi$ in seconds. 


We adopt the 3D Gaussians splatting (3DGS)~\cite{kerbl3Dgaussians} as the representation, which allows for photorealistic, real-time rendering and efficient pose control.
Each 3D Gaussian primitive is parameterized by its center location $\mathbf{p} \in \mathbb{R}^3$, directional scales $\boldsymbol{\sigma} \in \mathbb{R}^3$, and orientation (represented as a quaternion) $\mathbf{r} \in \mathbb{R}^4$. In addition, the primitive includes opacity $\rho \in [0,1]$ and spherical harmonic (SH) coefficients $\mathbf{f}$  to model view-dependent appearance.

Inspired by LHM~\cite{qiu2025LHM}, we employ a set of spatial points $P \in \mathbb{R}^{N_{\text{points}} \times 3}$ uniformly sampled from the SMPL-X surface in its canonical pose to serve as the anchors.
Conditioned on the multi-image inputs, these points are processed and decoded to regress the human 3D Gaussian appearance in canonical space through a feed-forward transformer-based architecture. The pipeline can be formulated as:
\begin{equation}
\begin{aligned}
    \mathbf{\chi}\{\mathbf{p}, \mathbf{r}, \mathbf{f}, \rho, \boldsymbol{\sigma\}} = \text{LHM++}(P \mid I^1,\dots,I^{N}).
\end{aligned}
\end{equation}
\paragraph{Model Design} 
A straightforward solution to this problem is to extend LHM to support multiple image inputs by directly concatenating all available image tokens and performing attention operations between 3D point tokens and image tokens. However, this naive extension results in significant computational and memory overhead due to the quadratic complexity of self-attention operations with respect to the total number of tokens, i.e., $\mathcal{O}((N_\textrm{points} + N)^2)$.

To mitigate this issue, we explore strategies to reduce the number of geometric point tokens involved in attention. However, we empirically observe that simply reducing the number of point tokens significantly degrades reconstruction performance.
To address this trade-off, we propose an efficient \emph{Encoder-Decoder Point-Image Transformer Framework} to fuse image features with geometric point features, as illustrated in \fref{fig:pipeline}, which maintains reconstruction quality while reducing the attention footprint. 
Additionally, with an increase in input images, image tokens dominate the attention computation. Inspired by ToME~\cite{bolya2022tome,bolya2023tomesd}, we merge the image tokens in a fixed ratio \(r\) based on the key similarity map calculated from frame-wise attention before global attention, then unmerge them after global attention. Benefiting from these designs, our model significantly reduces the memory consumption and inference latency and achieves stable performance gains compared to a straightforward extension of LHM.

The final geometric point features output from the decoder are utilized to regress 3D Gaussian parameters using lightweight multi-layer perceptron (MLP) heads. Subsequently, Linear Blend Skinning (LBS) is employed to animate the canonical avatar into the target pose. To enhance the quality of the animation rendering results, we develop a lightweight, real-time neural rendering network that improves rendering outcomes for novel views and poses based on 3D-aware human animation features.

\subsection{Encoder-Decoder Point-Image Transformer Framework}

To efficiently fuse multi-image features with 3D geometric information, we propose an encoder-decoder architecture based on \emph{Point-Image Transformer blocks (PIT-blocks)}. 
This framework enables hierarchical feature interaction while alleviating the computational and memory burden associated with dense attention.

We begin by projecting the SMPL-X anchor points in canonical space into a set of geometric tokens and encoding the input images into image tokens, as described in \sref{sec:tokenization}.
The encoder-decoder network consists of $N_\text{layer}$ PIT blocks. In the first $\lfloor N_\text{layer}/2 \rfloor$ encoder blocks, we progressively reduce the spatial resolution of the geometric tokens using Grid Pooling~\cite{wu2022ptv2}. At each layer, the downsampled point tokens perform the attention operation with the image tokens, enabling compact geometric representations enriched with multi-image visual cues.

In the subsequent $\lceil N_\text{layer}/2 \rceil$ decoder blocks, we upsample the geometric tokens to restore their original resolution. At each stage, the upsampled tokens are concatenated with the corresponding high-resolution features from the encoder via skip connections. These fused features are further refined by the PIT blocks to reconstruct detailed geometry and view-dependent appearance.

\subsection{Geometric Point and Images Tokenization}
\label{sec:tokenization}

\paragraph{Geometric Point Tokenization}
To incorporate human body priors, we initialize a set of 3D query points \(\mathbf{X} = \left\{ \mathbf{x}_i \right\}_{i=1}^{N_{\text{points}}} \subset \mathbb{R}^3\) by uniformly sampling from the mesh of a canonical SMPL-X pose. 
Following the design of Point Transformer v3 (PTv3)~\cite{wu2024ppt}, we first serialize these points into a structured sequence and then project them into a higher-dimensional feature space using an MLP. Formally, this process is expressed as:
\begin{equation}
\begin{aligned}
    X &= \text{Serialization}(X), \\
    \mathbf{T}_{\text{3D}} &= \text{MLP}_{\text{proj}}(X) \in \mathbb{R}^{N_{\text{points}} \times C_{\text{point}}},
\end{aligned}
\end{equation}
where \(C_{\text{point}}\) denotes the dimensionality of the point tokens.

\paragraph{Multi-Image Tokenization}
To obtain rich image features, we adopt DINOv2~\cite{oquab2023dinov2}, a vision transformer pretrained on large-scale in-the-wild datasets, as the image encoder \(\mathcal{E}_{\text{Img}}\). Given an input image \(I\), we extract a sequence of image tokens as follows:
\begin{equation}
    \mathbf{T}_{\text{I}} = \mathcal{E}_{\text{Img}}(I) \in \mathbb{R}^{N_{\text{I}} \times C},
\end{equation}
where \(N_{\text{I}}\) is the number of image tokens and \(C\) is the output feature dimension of the transformer.

\subsection{Point-Image Transformer Block}

After obtaining both geometric and image tokens, we design an efficient \emph{Point-Image Transformer Block} (PIT-block), which comprises three core attention modules to facilitate cross-modal interaction:

\paragraph{Point-wise Attention}
To model self-attention among geometric tokens, we adopt the patch-based point transformer blocks from PTv3~\cite{wu2024ppt}. This design enables cross-patch interactions via randomized shuffling of point orders, as detailed in the Supplementary Materials:
\begin{equation}
\mathbf{T}_{\text{3D}} = \text{PTv3-Block}(\mathbf{T}_{\text{3D}}).
\end{equation}

\paragraph{Image-wise Attention}
Given the image token sequence \(\mathbf{T}_{\text{2D}} = \{\mathbf{T}^1_{\text{I}}, \dots, \mathbf{T}^N_{\text{I}}\} \in \mathbb{R}^{N \times N_{\text{I}} \times C}\), we apply self-attention independently to the tokens of each image. This updates the features within each frame based on its own image tokens:
\begin{equation}
\mathbf{T}_{\text{2D}},\mathbf{K}_{\text{2D}} = \text{Self-Attention}(\mathbf{T}_{\text{2D}}).
\end{equation}
where \(\mathbf{K}_{\text{2D}} \in \mathbb{R}^{N N_{\text{I}} \times C_{\text{attn}}}\) represents the key features with \(C_{\text{attn}}\) channels from \textbf{QKV} in self-attention block, which are used to compute the similarity map of the image tokens.

\paragraph{Point-Image Attention}
After obtaining the updated features for both point-wise and frame-wise modalities, we develop a global point-image attention mechanism to fuse the point and multi-image tokens. Our model builds upon the powerful Multimodal-Transformer (MM-Transformer)~\cite{esser2024scaling} to efficiently merge features from different modalities. 

To enhance global context representation in the input images, we utilize the class token \(\mathbf{T}_{\text{cls}}\) extracted from the first frame as learnable global context features. 
Additionally, to align the dimensions of different modalities, we incorporate projection MLPs into both the input and output layers of the MM-Transformer~(MM-T):

\begin{equation}
\label{eq:mmt}
\begin{aligned}
    \mathbf{\bar{T}}_{\text{3D}}, \mathbf{\bar{T}}_{\text{2D}} &= \text{MLP}_{\text{proj}}(\mathbf{T}_{\text{3D}}) , \text{Merge}(\text{Flatten}(\mathbf{T}_{\text{2D}}),\mathbf{K}_{\text{2D}}),
    \\
    \mathbf{\bar{T}}_{\text{3D}}, \mathbf{\bar{T}}_{\text{2D}} &= \text{MM-T}(\mathbf{\bar{T}}_{\text{3D}}, \mathbf{\bar{T}}_{\text{2D}} \mid \mathbf{T}_{\text{cls}}), \\
    \mathbf{T}_{\text{3D}}, \mathbf{T}_{\text{2D}} &= \text{MLP}_{\text{uproj}}(\mathbf{\bar{T}}_{\text{3D}}) , \text{UnFlatten}(\text{UnMerge}(\mathbf{\bar{T}}_{\text{2D}},\mathbf{K}_{\text{2D}}), N). \\
\end{aligned}
\end{equation}


Specifically, $\text{MLP}_{\text{proj}}$ maps point tokens $\mathbf{T}_{\text{3D}} \in \mathbb{R}^{\bar{N}_{\text{points}} \times C_{\text{point}}}$ to $\mathbf{\bar{T}}_{\text{3D}} \in \mathbb{R}^{\bar{N}_{\text{points}} \times C}$, and $\text{MLP}_{\text{uproj}}$ performs the inverse mapping. The Merge and UnMerge operations are adapted from ~\cite{bolya2023tomesd}, with a merge ratio $r = 0.5$, reducing the number of image tokens from $NN_{\text{I}}$ to $NN_{\text{I}} / r$. This design reduces the time complexity of global attention from $\mathcal{O}((N_{\text{points}} + N)^2)$ to $\mathcal{O}((\bar{N}_{\text{points}} + N/r)^2)$.

\subsection{3D Human Gaussian Parameter Prediction}

Given the fused point tokens $\mathbf{T}_\text{3D}$ obtained from the encoder-decoder transformer framework, we predict the parameters of 3D Gaussians in the canonical human space using a lightweight MLP head:
\begin{equation}
\begin{aligned}
\label{eq:regress}
    \{\Delta\mathbf{p}_i, \mathbf{r}_i, \mathbf{f}_i, \rho_i, \boldsymbol{\sigma}_i\} &= \text{MLP}_{\text{regress}}(\mathbf{T}_{\text{3D}}^{(i)}), \\
    \mathbf{p}_i &= \mathbf{x}_i + \Delta\mathbf{p}_i, \quad \forall i \in \{1,\dots,N_{\text{points}}\},
\end{aligned}
\end{equation}
where \(\Delta\mathbf{p}_i \in \mathbb{R}^3\) denotes the predicted residual offset from the corresponding canonical SMPL-X vertex \(\mathbf{x}_i\), and \(\mathbf{r}_i, \mathbf{f}_i, \rho_i, \boldsymbol{\sigma}_i\) are the Gaussian orientation, feature vector, opacity, and scale, respectively.

\subsection{3D-Aware Human Animation Rendering}
The final output of our model is to drive the reconstructed avatar into view space using the provided human motion control signal, specifically the SMPL-X parameters. Since our Gaussian model is deformed based on SMPL-X, it struggles to accurately represent loose clothing, such as skirts. To improve the rendering quality, we develop a lightweight, 3D-aware DPT head for generating final animation outcomes, inspired by LVSM~\cite{jin2024lvsm} and HumanRAM~\cite{yu2025humanram}. Unlike HumanRAM, which requires processing through transformers with multi-view images for each frame, making it computationally expensive. In contrast, our approach allows for real-time generation. 

The neural refinement process can be divided into two parts:

\paragraph{Human Animation Feature Rasterization} As we construct the canonical avatar in canonical space, we can leverage 3D priors to provide fine-grained control information. Specifically, given a reconstructed avatar obtained from Eq.~ \ref{eq:regress} and multi-scale geometric tokens $\mathbf{T}_{\text{3D}}$ from Eq.~\ref{eq:mmt}, we employ Gaussian Rasterization~\cite{qian20243dgs}, denoted as $\text{F}_{\text{GS}}$, to render high-dimensional geometric features into view space, resulting in $\mathbf{I}_{\text{feat}} \in \mathbb{R}^{H \times W \times C_{\textbf{point}}}$. Notably, as no 3D points are available to model the background, we utilize learnable background parameters $\mathbf{F}_{\text{back}}$, to represent background colors. The entire process can be formulated as follows:
\begin{equation}
\label{eq:fgs}
\mathbf{I}^{l}_{\text{feat}} = \text{F}_{\text{GS}}(\mathbf{T}^{l}_{\text{3D}}, \mathbf{r}, \mathbf{p}, \rho, \boldsymbol{\sigma}; \mathbf{F}_{\text{back}}),
\end{equation}
where $\mathbf{T}^{l}_{\text{3D}}$ represents the geometry features extracted from the $l$-th decoder layer of the PI Transformer framework.

\paragraph{3D-Aware Neural Animation Rendering} After obtaining the multi-scale 3D-aware human animation feature map $\mathbf{I}^{l}_{\text{feat}}$, we employ a lightweight neural network, as proposed in~\cite{ranftl2021dpt}, to fuse these multi-scale animation features for novel view and pose rendering:
\begin{equation}
\begin{aligned}
\label{eq:dpt}
    \mathbf{T}^{l}_{\text{feat}} &= \text{Patchify}(\mathbf{I}^{l}_{\text{feat}}), \\
    \hat{I}_{\text{NR}},\hat{M}_{\text{NR}}  &= \text{DPT-head}(\mathbf{T}^{0}_{\text{feat}}, \ldots, \mathbf{T}^{{N_\text{layer}/2}}_{\text{feat}}), \\
\end{aligned}
\end{equation}
where $\text{Patchify}$ refers to the operation that converts the 3D-aware animation features into image tokens using $8 \times 8$ patch windows, $\hat{I}_{\text{NR}}$ and $\hat{M}_{\text{NR}}$ respectively representing the final predicted RGB image and mask results.

\subsection{Loss Function}

\label{sec:target_function}

Our training strategy integrates photometric supervision from unconstrained video sequences with geometric regularization on Gaussian primitives. This hybrid optimization framework enables the learning of deformable human avatars without the need for explicit 3D ground-truth annotations.

To better capture complex clothing deformations, we adopt a diffused voxel skinning approach as proposed in~\cite{lin2022fite, qiu2023recmv}. Given the predicted 3DGS parameters, we transform the canonical avatar into target view space using voxel-based skinning.

\paragraph{Photometric Loss}
We render the animated Gaussian primitives via differentiable splatting to obtain an RGB image \(\hat{I}\) and an alpha mask \(\hat{M}\), based on the target camera parameters. Additionally, we generate the neural rendering results \(\hat{I}_{\text{NR}}\) and \(\hat{M}_{\text{NR}}\) using a lightweight DPT head. Supervision is applied through the following photometric loss:
\begin{equation}
    \mathcal{L}_{\text{photometric}} = \lambda_{\text{rgb}} \mathcal{L}_{\text{color}} + \lambda_{\text{mask}} \mathcal{L}_{\text{mask}} + \lambda_{\text{per}} \mathcal{L}_{\text{lpips}},
\end{equation}
where \(\mathcal{L}_{\text{color}}\) and \(\mathcal{L}_{\text{mask}}\) represent L1 losses on the RGB and alpha values, corresponding to the 3D Gaussian splatting and neural rendering losses, respectively. The term \(\mathcal{L}_{\text{lpips}}\) denotes a perceptual loss measures how similar the high-frequency features of neural-rendered and Gaussian Splat–rendered images are to those of the ground-truth images. The weights are set to \(\lambda_{\text{rgb}} = 1.0\), \(\lambda_{\text{mask}} = 0.5\), and \(\lambda_{\text{per}} = 1.0\).

\paragraph{Gaussian Regularization Loss}
Although the photometric loss offers effective supervision in the target view space, the canonical representation remains under-constrained due to the ill-posed nature of monocular reconstruction. This limitation leads to deformation artifacts when warping the avatar into novel poses. To address this fundamental challenge, we employ Gaussian regularization loss \(\mathcal{L}_{\text{reg}}\) to improve the learning of unobserved parts (see supplementary material for details).

\paragraph{Overall Loss}
The overall training objective combines photometric reconstruction accuracy with geometric regularization:
\begin{equation}
    \mathcal{L}_{\text{total}} = \mathcal{L}_{\text{photometric}} + \mathcal{L}_{\text{reg}}.
\end{equation}

\section{Experiments}
\label{sec:experiments}

\begin{table*}[tb]\centering
    \caption{Comparison with sparse-view input methods. The results are averaged on the public and smartphone-captured video sequences.}
    \label{tab:animation_results_combined}

\resizebox{0.96\textwidth}{!}{
\large
\begin{tabular}{c| cccc | cccc | cccc | cccc}
\toprule
\multirow{2}{*}{\textbf{views}} & 
\multicolumn{4}{c|}{\textbf{InstantAvatar}~\cite{jiang2023instantavatar}} & 
\multicolumn{4}{c|}{\textbf{GaussianAvatar}~\cite{hu2024gaussianavatar}} & 
\multicolumn{4}{c|}{\textbf{ExAvatar}~\cite{moon2024exavatar}} & 
\multicolumn{4}{c}{\textbf{LHM++}} \\
& PSNR & SSIM & LPIPS & Time & PSNR & SSIM & LPIPS & Time & PSNR & SSIM & LPIPS & Time & PSNR & SSIM & LPIPS & Time \\
\midrule
2  & 23.337 & 0.944 & 0.091 &  {3.6~min} & 22.008 & 0.947 & 0.067 & 3.8~min &  26.588          & 0.965 & 0.036 & 8.5~min &  \textbf{27.835} & \textbf{0.970} & \textbf{0.018} &  \textbf{1.08~sec} \\
4  & 23.391 & 0.943 & 0.093 & 6.0~min & 23.382 & 0.952 & 0.051 & 5.6~min & 27.293  & 0.966 & 0.034 & 15~min &  \textbf{27.940} & \textbf{0.971} & \textbf{0.018} &  \textbf{1.40~sec} \\
8  & 23.650 & 0.943 & 0.090 & 10.6~min & 24.106 & 0.956 & 0.048 &  {9.0~min} & 28.006         & 0.968 & 0.031 & 32~min &  \textbf{28.147} & \textbf{0.971} & \textbf{0.017} &  \textbf{2.09~sec} \\
16 & 24.235 & 0.949 & 0.075 & 18.8~min & 24.398 & 0.957 & 0.047 &  {15.7~min} &  28.358          & 0.969 & 0.030 & 72~min &  \textbf{28.394} & \textbf{0.972} & \textbf{0.016} &  \textbf{3.66~sec} \\
64 &  25.271 & 0.956 & 0.055 & 54.0~min & 25.308 & 0.960 & 0.040 & 66~min & \textbf{29.587} & 0.971 & 0.027 & 264~min & 28.422          & \textbf{0.972} & \textbf{0.016} & \textbf{17.94~sec}\\
\bottomrule
\end{tabular}
}
    \vspace{-1em}
\end{table*}

We design three variants of our model with \(N_{\text{layer}} = 4, 6, 8\) layers of the PI-MT block, corresponding to \MethodName (S), \MethodName (M), and \MethodName (L), respectively. The models comprise approximately 500M, 700M, and 900M training parameters in total. The training details can be found in the supplementary material. For brevity, \MethodName indicates \MethodName (M) unless otherwise specified.

\paragraph{Evaluation Protocol} We report PSNR, SSIM~\cite{wang2003multiscale}, and LPIPS~\cite{zhang2018unreasonable} to assess rendering quality. If not specifically written, and measure efficiency with GPU memory usage and inference time on a single NVIDIA A100-80G. For our evaluation benchmarks, we establish three datasets: public animation videos, smartphone-captured videos for sparse view reconstruction, and in-the-wild fashion videos for both single-view and sparse view settings. All evaluation datasets will be made publicly available for future research.

\subsection{Comparison with Existing Methods}
\paragraph{Animatable Human Reconstruction from Sparse Images} 
We first evaluate our method against three state-of-the-art fitting-based approaches designed for sparse inputs: InstantAvatar~\cite{jiang2023instantavatar}, GaussianAvatar~\cite{hu2024gaussianavatar}, and ExAvatar~\cite{moon2024exavatar}. For this evaluation, we utilize two datasets: (1) a public animation dataset comprising 23 video sequences from NeuMan~\cite{jiang2022neuman}, REC-MV~\cite{qiu2023recmv}, and Vid2Avatar~\cite{guo2023vid2avatar}; and (2) a smartphone-captured video dataset containing 20 clips collected from in-the-wild environments.



\begin{figure*}[tb] 
    \centering
    \includegraphics[width=1.00\textwidth]{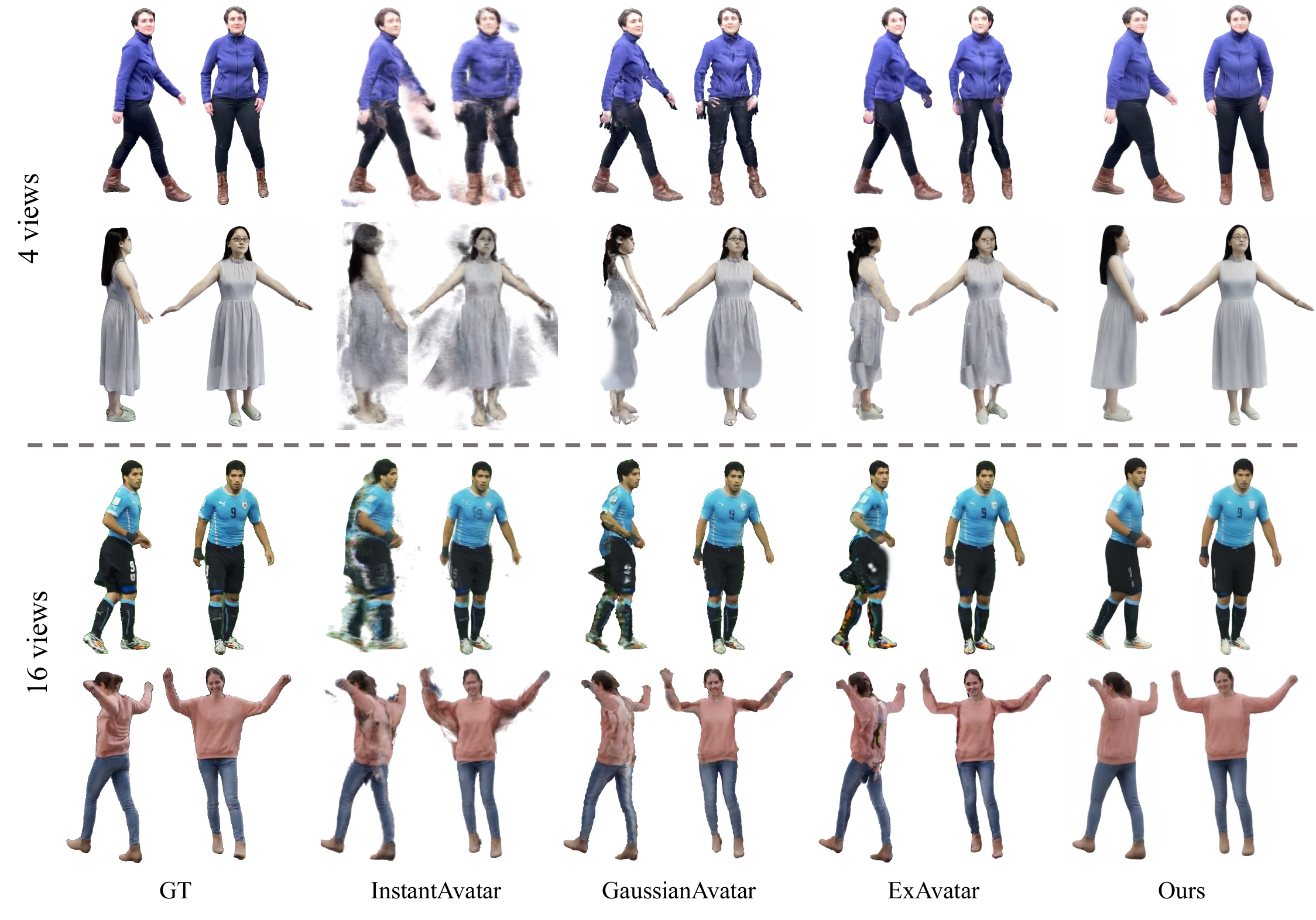} \\
    \caption{Animatable human reconstruction comparisons from sparse images.} 
    \label{fig:animation_in_the_wild}
    \vspace{-1em}
\end{figure*}

\begin{table}[tb] 
    \centering
    \caption{Comparison with single-image methods on pose animation on our in-the-wild fashion video dataset.}
    \label{tab:animation_singleview}
    

\resizebox{0.48\textwidth}{!}{
\large
\begin{tabular}{l*{8}{c}}
    \toprule
   Methods & Input & PSNR $\uparrow$ & SSIM $\uparrow$ & LPIPS $\downarrow$ & Time$\downarrow$ & Memory $\downarrow$ \\
    \midrule
    AniGS~\cite{qiu2024AniGS} & 1 & 17.234 & 0.812 & 0.103  & 900.00~s & 24.00~GB \\
    IDOL~\cite{zhuang2024idol}& 1 & 17.912 & 0.847  & 0.097  & 1.93~s & 23.12~GB \\
    \midrule
    \multirow{5}{*}{LHM-0.7B~\cite{qiu2025LHM}} & 1  & 21.233 & 0.877 & 0.078 & 5.73~s & 19.59~GB \\
    & 8 & 21.761 &0.883 & 0.073 & 24.76~s & 25.42~GB \\
    & 16 & 21.868 & 0.885& 0.069 & 49.55~s & 28.77~GB\\
    & 64 & 21.852 & 0.885 & 0.069& 381.48~s& 51.73~GB\\
    
    \midrule
    \multirow{5}{*}{LHM++} & 1 & 21.518 &  0.882  & 0.068 & 0.91 s & 6.78~GB\\
    & 8  & 22.208 & 0.886 & 0.060& 2.09~s & 7.35~GB\\
    & 16 & \Scnd{22.354} & \Frst{0.887} & \Frst{0.057} & 3.66~s & 8.48~GB\\
    & 64 & \Frst{22.363} & \Scnd{0.887} & \Scnd{0.057} & 17.94~s & 29.51~GB \\
    \bottomrule
\end{tabular}
}
    \vspace{-4mm}
\end{table}

\Tref{tab:animation_results_combined} summarizes quantitative experiments evaluating our model against baseline methods on public and smartphone-captured video datasets. 
Specifically, compared to ExAvatar, our method generates animatable avatars within seconds, whereas ExAvatar requires 15 minutes to over an hour. 
Additionally, unlike ExAvatar, which relies on dozens of input images, our model achieves higher accuracy with far fewer inputs. 
As illustrated in \Fref{fig:animation_in_the_wild}, while \MethodName marginally outperforms ExAvatar with 16-view inputs, ExAvatar produces severe artifacts such as geometric distortions and texture blurring. 
In contrast, our method yields vivid and realistic animation results. 

\begin{figure*}[tb]
    \centering
    \includegraphics[width=0.9\textwidth]{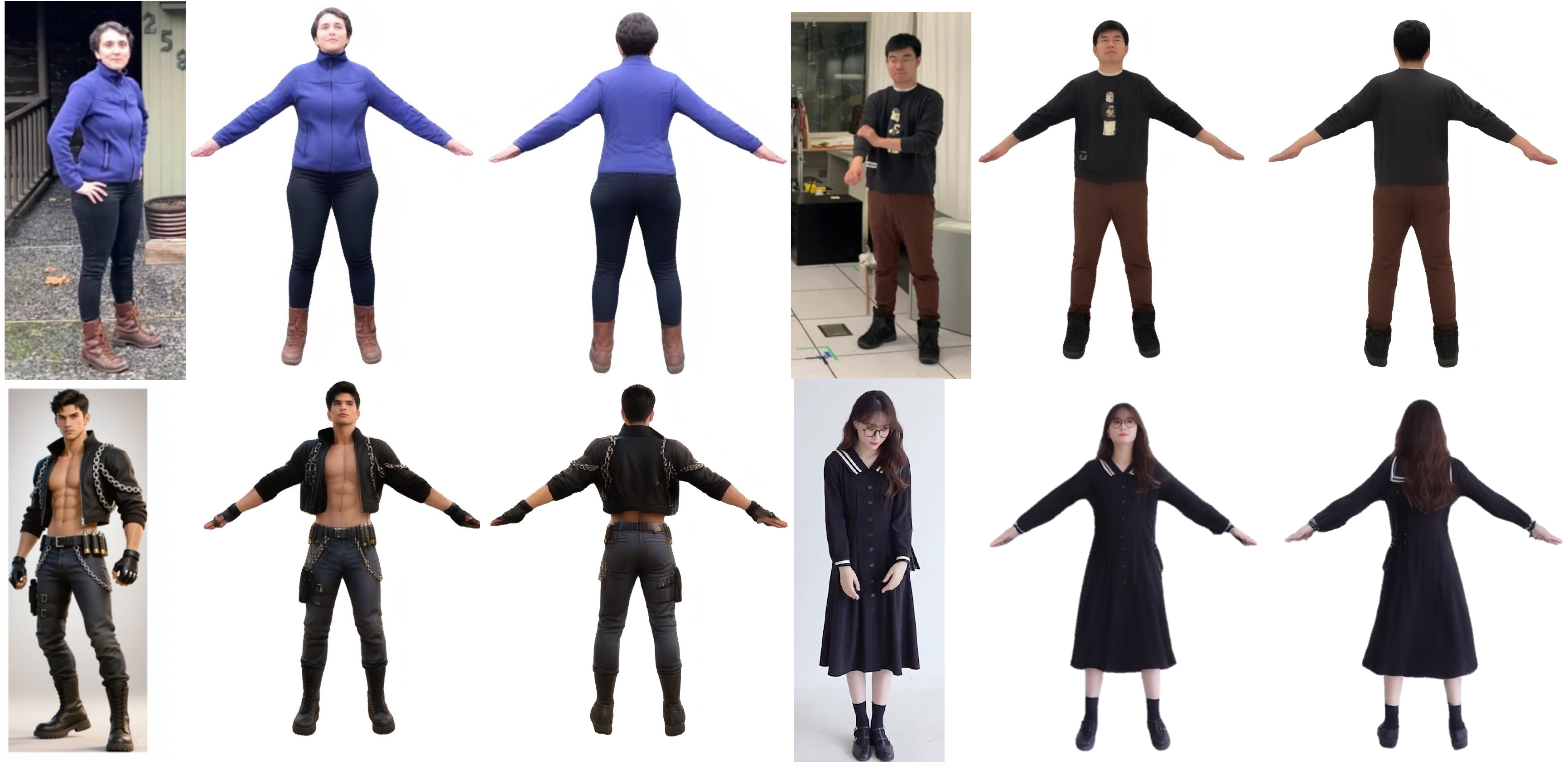}
    \caption{Human reconstruction of \MethodName in canonical space from sparse image inputs.}
    \label{fig:tpose_reconstruction}
\end{figure*}

\Fref{fig:tpose_reconstruction} shows the avatar reconstruction results of \MethodName in canonical space from sparse images, demonstrating the robustness of the proposed model across diverse input categories, such as real and cartoon-style inputs.

\begin{figure*}[tb]\centering
\includegraphics[width=0.87\textwidth]{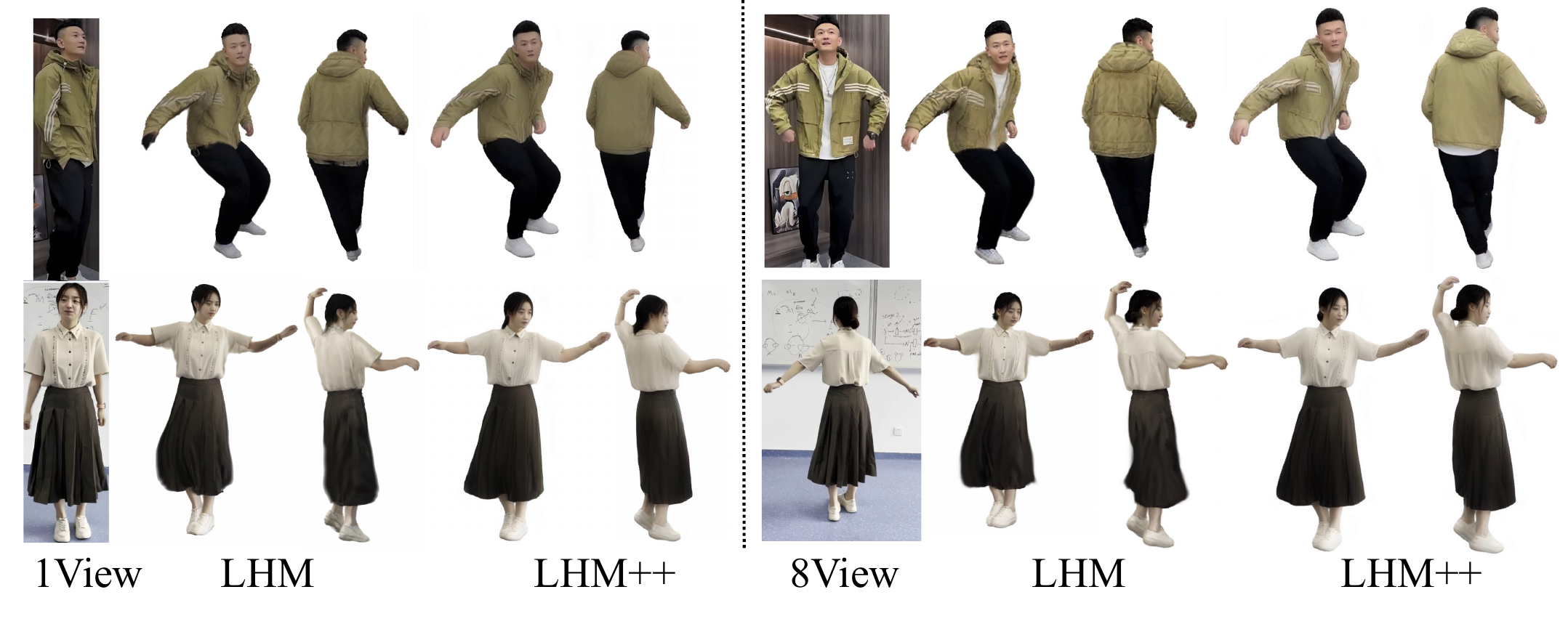}
\caption{Qualitative comparison with LHM. \MethodName matches LHM's quality with a single input view and generates progressively more detailed results as view count increases. Please zoom in for better view.}
\label{fig:animation_comparison_with_single_view}
\end{figure*}

\paragraph{Animatable Human Reconstruction from a
Single Image} To assess performance under extreme sparse conditions, single-view input, we compare \MethodName against two state-of-the-art single-image methods: AniGS~\cite{qiu2024AniGS} and IDOL~\cite{zhuang2024idol}. 
We use 400 in-the-wild fashion videos as the evaluation benchmark. 
As demonstrated in \Tref{tab:animation_singleview}, \MethodName outperforms AniGS and IDOL by $4.28$ and $3.61$ dB in PSNR, respectively. 
In terms of LPIPS, we observe improvements of $0.35$ and $0.29$ against the same baselines, validating the superiority of our approach in single-view settings.

\paragraph{Animatable Human Reconstruction from Any Images} We also compare against LHM~\cite{qiu2025LHM}, which was originally trained on single-view data. 
To enable it to handle arbitrary-view inputs, we align its configuration with our method and fine-tune it for any-view inputs. Notably, we retain LHM's original configuration of 40K query points, as increasing this number to match our method resulted in out-of-memory errors. 
As shown in \Tref{tab:animation_singleview}, \MethodName consistently outperforms LHM across 1-, 8-, 16-, and 64-view inputs, achieving PSNR gains of $0.285$, $0.447$, $0.486$, and $0.511$ dB, respectively. 
Beyond accuracy, our method achieves approximately $10\times$ faster inference speed while consuming only a quarter of the memory compared to LHM. 
\Fref{fig:animation_comparison_with_single_view} presents a qualitative comparison using 1- and 8-view inputs, where our method produces significantly more realistic and detailed results, demonstrating robust scalability from single to dense views.

Additional qualitative and quantitative results can be found in the appendix.

\begin{table*}[tb]
    \caption{Effectiveness of the Encoder-Decoder PIT under varying inputs and points.}
    \label{tab:efficiency_analysis}
    \resizebox{0.8\textwidth}{!}{
\large
\begin{tabular}{l |cc | cc | cc | cc}
\toprule
\multirow{2}{*}{\textbf{\# Points}} & 
\multicolumn{2}{c|}{\textbf{1 view}} & 
\multicolumn{2}{c|}{\textbf{4 views}} & 
\multicolumn{2}{c|}{\textbf{8 views}} & 
\multicolumn{2}{c}{\textbf{16 views}} \\
& LHM-0.7B & LHM++ & LHM-0.7B & LHM++ & LHM-0.7B & LHM++ & LHM-0.7B & LHM++ \\
\midrule
40~K  &  5.16~s & \textbf{0.55~s} & 9.41~s & \textbf{0.74~s} & 15.23~s & \textbf{1.01~s} & 32.32~s & \textbf{1.74~s} \\
80~K  &  19.47~s & \textbf{0.69~s} & 25.73~s & \textbf{0.88~s} & 35.36~s & \textbf{1.19~s} & 59.25~s & \textbf{1.95~s}  \\
160~K &  74.31~s & \textbf{0.79~s} & 85.58~s &\textbf{1.00~s} & 102.44~s & \textbf{1.31~s} & 140.67~s & \textbf{2.13~s} \\

\bottomrule
\end{tabular}
}

    \centering
\end{table*}

\subsection{Ablation Study}
In this section, we present comprehensive ablation studies demonstrating that \MethodName is not a simple extension of LHM but a carefully designed, novel, and efficient encoder–decoder framework.

\begin{figure*}[htb] \centering
    \vspace{-1em}
    \includegraphics[width=0.95\textwidth]{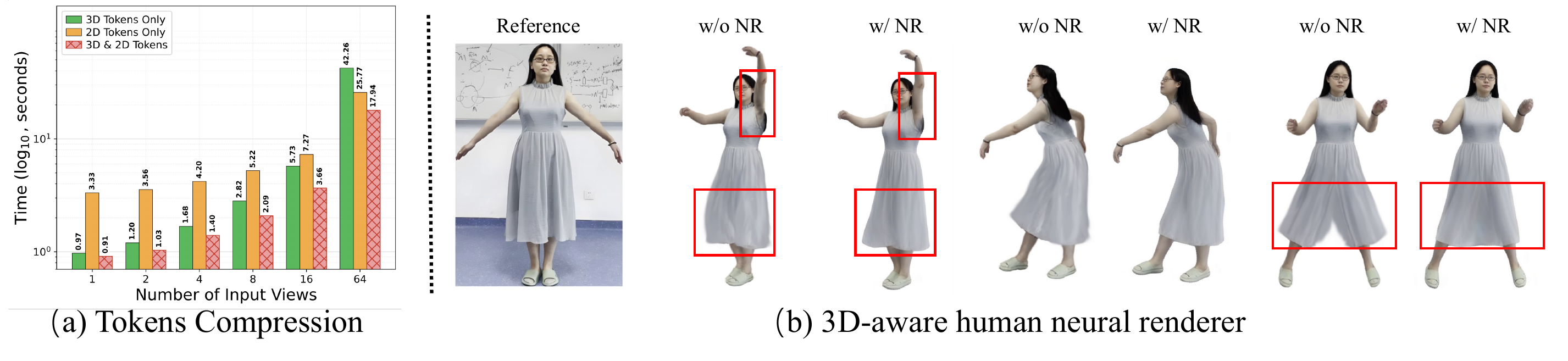}
    \caption{The ablation study for both (a)~tokens compression and (b)~3D-aware human neural renderer. `NR' is the abbreviation for `3D-aware neural renderer'. The \fcolorbox{red}{white}{red boxes} highlight the visual difference. Please zoom in for better view.} 
    \label{fig:ablation-for-3d-aware}
    \vspace{-1em}
\end{figure*}

\paragraph{Effectiveness of the Encoder-Decoder PIT} \Tref{tab:efficiency_analysis} quantifies the effectiveness of the encoder-decoder PIT. LHM processes all tokens directly in attention, resulting in quadratic complexity that scales poorly with increasing 3D token counts. In contrast, our encoder–decoder architecture compresses redundant tokens early in the pipeline, substantially reducing the overall computational load. At matched 3D point counts, PIT achieves 70--100$\times$ faster fusion than LHM, demonstrating the effectiveness of our token compression strategy.

\paragraph{Token Compression Across Image Numbers} Our framework accelerates attention process by compressing 3D geometric tokens via grid pooling and 2D image tokens through token merging~(details can be found in supplementary). To evaluate the efficacy of these strategies, \fref{fig:ablation-for-3d-aware}(a) analyzes the impact of token compression across varying numbers of input views. Under sparse-view settings, geometric tokens dominate inference cost; compressing them yields major speedups. As the number of input image increases, image tokens become predominant, shifting the computational bottleneck to image token processing. By jointly compressing both token types, our method ensures consistent efficiency gains across both sparse and dense input configurations.

\begin{table}[tb] 
    \centering
    \caption{Comparison with LHM for multi-modality fusion on  our in-the-wild fashion dataset with 8 input views.}
        \label{tab:multi_fusion_for_nums}
    \resizebox{0.48\textwidth}{!}{
\large
\begin{tabular}{l | cc | cc | cc}
\toprule
\multirow{2}{*}{\textbf{Methods}} & 
\multicolumn{2}{c|}{\textbf{40~K}} & 
\multicolumn{2}{c|}{\textbf{80~K}} & 
\multicolumn{2}{c}{\textbf{160~K}} \\
& PSNR & Time & PSNR & Time & PSNR & Time   \\
\midrule
LHM-0.7B  & \textbf{21.761} & 24.76~s & 21.803  & 49.78~s & 21.796 & 110.37~s \\
\MethodName &  21.747 & \textbf{1.86~s} & \textbf{22.124} & \textbf{1.98~s} & \textbf{22.208} & \textbf{2.09~s} \\
\bottomrule
\end{tabular}
}

\end{table}

\paragraph{Effect of 3D Query Token Count} Building on the analysis presented above, we next evaluate the performance improvements of the proposed method over LHM. First, we study how the number of 3D query tokens influences both methods. \Tref{tab:multi_fusion_for_nums} reports comparative experiments on the in-the-wild fashion dataset with eight input views, varying the number of sampling points. The results indicate that LHM gains only marginally as the count of 3D tokens increases. We attribute this to LHM’s inefficient fusion mechanism. LHM's architecture learns very large attention maps, so redundant tokens impede learning and degrade fusion efficiency. By contrast, the encoder–decoder PIT design effectively prunes redundant 3D and 2D tokens, substantially improving the model’s fusion performance.

\paragraph{Effect of Attention Modules}
We conduct an ablation study to investigate the impact of attention modules within the encoder–decoder PIT architecture on multi-modal fusion across in-the-wild fashion, as well as public animation and  smartphone-captured video datasets~(Public \& Smartphone).  As detailed in \Tref{tab:ablation_attention_module}, incorporating image-wise attention enhances per-frame feature representation compared to relying solely on multi-modal attention. 
Furthermore, point-wise attention facilitates the fusion of local geometric features by leveraging Euclidean distances within the point cloud. This mechanism promotes effective point-to-point context exchange, also improving overall model performance. Finally, the synergistic combination of both attention modules yields the most significant performance gains.

\begin{table}[tb] 
    \caption{effect of attention modules in the encoder–decoder PIT.}
        \label{tab:ablation_attention_module}
\resizebox{0.48\textwidth}{!}{
\large
\begin{tabular}{l | cc | cc }
\toprule
\multirow{2}{*}{\textbf{Methods}} & 
\multicolumn{2}{c|}{\textbf{In-the-wild Fashion}} & 
\multicolumn{2}{c}{\textbf{Public \& Smartphone}} \\
& 4 Views & 16 Views & 4 Views & 16 Views   \\
\midrule
\MethodName w/o Image/Point-wise Attention & 21.669 & 21.977 & 27.403 &  27.655  \\
\MethodName w/o Image-wise Attention & 21.735 & 22.136  & 27.656 & 27.927   \\
\MethodName  w/o Point-wise Attention  & \Scnd{21.838} & \Scnd{22.230}  &  \Scnd{27.764} & \Scnd{28.129}  \\
\MethodName & \Frst{21.957} & \Frst{22.354} & \Frst{27.940} & \Frst{28.394} \\

\bottomrule
\end{tabular}
}

    \centering
\end{table}

\paragraph{Effect of 3D-Aware Human Neural Renderer} \Fref{fig:ablation-for-3d-aware}(b) illustrates the qualitative results of the ablation study for 3D-aware neural renderer. Since our \MethodName directly learns a positional residual from the SMPL-X template, challenges remain in rendering loose-fitting clothing, despite employing mask distribution loss. The proposed neural renderer not only effectively addresses this issue, but also significantly enhance the overall rendering results.

Additional ablation studies are provided in the appendix, including scalability, effect of image token merging and the detailed analysis on neural renderer.

\section{Conclusion}
\label{sec:Conclusion}

We present \MethodName, an efficient feed-forward framework for rapid and high-fidelity 3D human avatar reconstruction from one or a few casually captured, pose-free images. 
Our approach introduces the Encoder-Decoder Point-Image Transformer (PIT), which enables efficient multimodal fusion between geometric point tokens and sparse multi-view image tokens through hierarchical attention.
In addition, we introduce a lightweight 3D-aware neural animation renderer to refine the rendering quality of reconstructed avatars.
Extensive experiments across synthetic and real-world datasets demonstrate that \MethodName\ effectively unifies single- and sparse-view reconstruction and supports realistic avatar animation.



{
    \small
    \bibliographystyle{ieeenat_fullname}
    \bibliography{main}
}
\appendix
\clearpage

\section{Overview}
This supplementary material provides comprehensive details supporting the main paper.


We begin with \sref{sec:demo_video} (\textbf{Demo Video}), which presents animation results, comparative experiments, and a visual gallery across diverse datasets. 

Subsequently, \sref{sec:more_methods} (\textbf{Method Details and Analysis}) encompasses the Encoder-Decoder Point-Image Transformer architecture~(\sref{sec:ed_architecture}), image token merging strategies~(\sref{sec:merging}), and Gaussian regularization losses~(\sref{sec:gaussian_regular}). This section also provides a detailed analysis of the 3D-aware neural renderer~(\sref{sec:renderer}) and our implementation configurations~(\sref{sec:imple}).

Next, \sref{sec:experiments} (\textbf{Experiments}) provides comprehensive experimental details, including the data protocols employed from public animation datasets~(\sref{sec:data}). It presents comparisons with commercial image-to-3D methods (e.g., Hunyuan-2.5)~(\sref{hunyuan}) and extensive ablation studies on scalability~(\sref{sec:scalability}) and training/inference efficiency~(\sref{sec:efficiency}). Furthermore, we provide comprehensive quantitative~(\sref{sec:quan}) and qualitative results~(\sref{sec:qual}) on evaluation datasets to demonstrate the model's robustness and generalization capabilities. 

Finally, \sref{sec:limitation} discusses the limitations of \MethodName, presents representative failure cases, and proposes potential remedies.




\section{Demo Video}
\label{sec:demo_video}
Please refer to the \href{https://www.youtube.com/watch?v=Nipf3jdSi34}{\textcolor{linkred}{Demo Video}} for animation results of the reconstructed 3D avatars. 
Our demo consists of three components: (1) rendering results from varying input views, (2) comparison experiments against methods such as ExAvatar~\cite{moon2024exavatar}, LHM~\cite{qiu2025LHM}, and IDOL~\cite{zhuang2024idol}, and (3) an animated video gallery across public animation and smartphone-captured datasets, in-the-wild fashion videos, and AI-generated images.

\paragraph{Details of Testing Dataset in Demo} 
For comparison experiments with ExAvatar, we utilize 16-view images sampled uniformly from \text{00000\_random} and \text{00069\_Dance} in Vid2Avatar~\cite{guo2025vid2avatar}, \text{anran\_self\_rotated} from REC-MV~\cite{qiu2023recmv}, and \text{Bike} from NeuMan~\cite{jiang2022neuman}. 
For comparison experiments with feed-forward approaches, we use \text{Anran\_Skirt} from REC-MV~\cite{qiu2023recmv} and two in-the-wild monocular video samples.

Next, we provide animated gallery videos to evaluate our model's generalization capability across four public datasets. 
For REC-MV~\cite{qiu2023recmv}, we select \text{anran\_purple} and \text{xiaolin}. 
For NeuMan~\cite{jiang2022neuman}, we use clips from \text{bike}, \text{citron}, and \text{jogging}. 
For Vid2Avatar~\cite{guo2025vid2avatar}, we choose \text{Yuliang}, \text{00000\_random}, \text{exstrimalik}, and \text{Suarez}. 
Lastly, for MVHumanNet~\cite{xiong2024mvhumannet}, we utilize samples with IDs \text{101157}, \text{103383}, \text{103528}, and \text{104316}.

\paragraph{Details of In-the-Wild Fashion Videos} 
For in-the-wild fashion videos, we utilize web-sourced monocular videos and randomly sample 8 views from the original clips. 

\paragraph{Details of AI-Generated Images} 
For AI-generated images, we employ FLUX.1 Kontext~\cite{labs2025flux1kontextflowmatching} to generate four different viewpoint images using prompts such as ``the front view of'', ``the back view of'', ``the right view of'', and ``the left view of''.

\section{More Details of Method and Analysis}
\label{sec:more_methods}

\subsection{Encoder-Decoder Point-Image Transformer}
\label{sec:ed_architecture}
\paragraph{Points Serialization}
To ensure both scalability and efficiency within our feed-forward framework, we employ serialization to convert unstructured SMPL-X anchor points into a structured format. Following Point Transformer v3 (PTv3)~\cite{wu2024ppt}, we utilize a mixture of four serialization patterns: Z-order, Hilbert, Trans Z-order, and Trans Hilbert. To prevent the model from overfitting to a specific point ordering, we randomly shuffle the serialization pattern sequence order to each PTv3 block.

\begin{figure}[htb] \centering
    \includegraphics[width=0.48\textwidth]{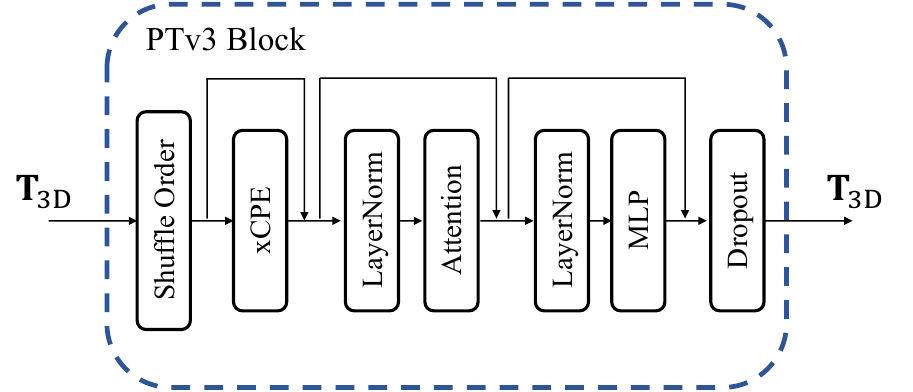}
    \caption{Detailed architecture of the PTv3 Block~\cite{wu2024ppt}.} \label{fig:ptv3_block}
\end{figure}

\paragraph{PTv3 Block}
As depicted in Fig.~\ref{fig:ptv3_block}, our point attention module adopts the architecture from Point Transformer v3~\cite{wu2024ppt}, leveraging patch-based self-attention to accelerate inference. Additionally, we incorporate Grid Pooling to downsample the point cloud, which significantly improves the computational efficiency of the self-attention mechanism. Regarding implementation details, patch sizes are configured to decrease progressively across downsampling stages. Specifically, we use patch sizes of \{4096, 2048, 1024\} for 80K point clouds and \{8192, 4096, 2048\} for 160K point clouds.


\begin{figure}[htb] \centering
    \includegraphics[width=0.48\textwidth]{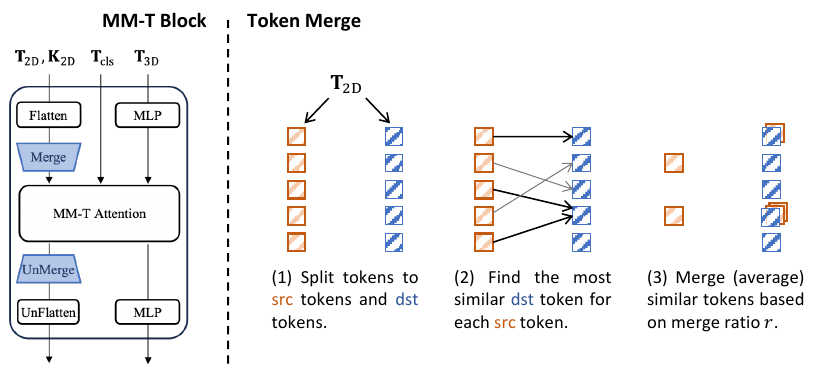}
    \caption{Illustration of the image token merging process. Given 2D tokens $\mathbf{T}_{\text{2D}}$ and key features $\mathbf{K}_{\text{2D}}$ from the preceding frame-wise attention block, source tokens are randomly sampled. A similarity matrix is then computed between the source and target tokens (derived from $\mathbf{K}_{\text{2D}}$), allowing the most similar source tokens to be greedily merged into their corresponding target tokens.}
    \label{fig: tome}
\end{figure}

\paragraph{3D Token Evolution}
Table~\ref{tab:ablation_3D_num} reports the evolution of 3D token counts across the six encoder layers of the PIT. To manage this reduction, we apply a point-cloud grid-pooling operator that efficiently compresses redundant tokens based on their Euclidean proximity.

\begin{table}[ht]
    \centering
    \caption{Number of 3D tokens across the encoder layers of the PIT.}
    \label{tab:ablation_3D_num}
    \resizebox{0.4\textwidth}{!}{
\large
\begin{tabular}{l*{8}{c}}
    \toprule
   Operator & 40K & 80K & 160K \\
    \midrule
    Grid Pooling~1 & 16778 & 22380 & 24377  \\
    Point-Image Attention~1& 16778 & 22380 & 24377  \\
    Grid Pooling~2 & 5547 & 6036 & 6441  \\
    Point-Image Attention~2& 5547 & 6036 & 6441  \\
    Grid Pooling~3& 1493 & 1625 & 1686  \\
    Point-Image Attention~3 & 1493 & 1625 & 1686  \\
    \bottomrule
\end{tabular}
}
\end{table}

\subsection{Image Token Merging}
\label{sec:merging}
As the number of input images increases, the image tokens gradually dominate the computing resources. Inspired by ToME~\cite{bolya2022tome, bolya2023tomesd}, we employ Bipartite Soft Matching to merge redundant image tokens, which helps to reduce inference times during the global attention process. As illustrated in \fref{fig: tome}, we construct a bipartite graph using stride sampling as proposed in ToME-SD~\cite{bolya2023tomesd}. To build the similarity matrix, we utilize the 'Key' tokens from the frame-wise attention mechanism described in the paper. Subsequently, we merge the similar image tokens based on a fixed merge ratio \(r\). The merged image tokens are then inputted to multi-modal attention with the geometric 3D tokens. After the multi-modal attention operation, we unmerge the image tokens back to their original size using a simple feature copying operator, allowing them to proceed to the next PI transformer block, and not change their original size.

\paragraph{The effect of Image Token Compression} Surprisingly, we found that reducing a certain proportion of image tokens does not degrade model performance. in contrast, it can even lead to a slight improvement. We hypothesize that the performance gain benefits from the removal of redundant tokens, making the attention map learning more effective. \Tref{tab:ablation_for_ratio} presents the details of the ablation study examining the ratio of token merging.

\begin{table}[ht]
    \centering
    \caption{\textbf{Effect of image token merging across reduction ratios} The metrics are calculated based on 16-view image inputs on in-the-wild fashion
    videos dataset.}
    \label{tab:ablation_for_ratio} 
    \resizebox{0.4\textwidth}{!}{
    \large
    \begin{tabular}{*{5}{c}}
        \toprule
        Methods & PSNR $\uparrow$ & SSIM $\uparrow$ & LPIPS $\downarrow$ & Time $\downarrow$ \\
        \midrule
        ratio=0.00 & 22.317 & 0.886 & 0.059 & 5.73~s\\
        ratio=0.25 & 22.352 & \textbf{0.887} & 0.058 & 4.82~s \\
        ratio=0.50 & \textbf{22.354 }& 0.887 & \textbf{0.057 } & 3.66~s\\
        ratio=0.75 & 22.335 & 0.886 & 0.059 & 3.07~s\\
        \bottomrule
    \end{tabular}
    } 
\end{table}

\paragraph{Analysis on Image Token Merging} To further analyze the `merge' and `unmerge' operator in global image-point attention, we conducted both qualitative and quantitative experiments that demonstrate how performance varies with different merge ratios and numbers of views.

\begin{figure*}[htb] \centering
    \includegraphics[width=0.96\textwidth]{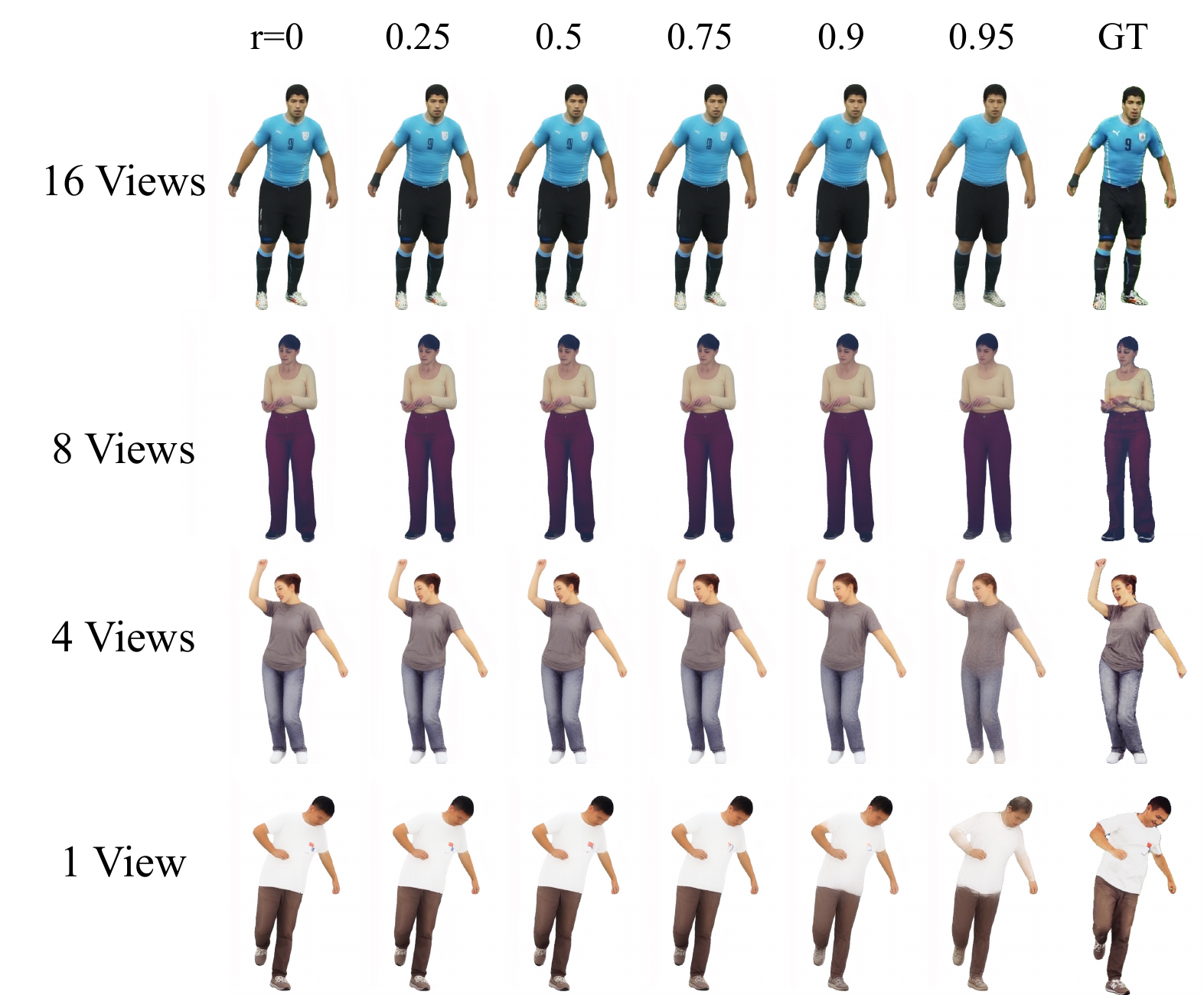} \\
    \caption{Ablation study of image-token merging. `r' denotes the merge ratio ($r=0$ means no merging). Larger `r' values correspond to greater compression of image tokens. Please zoom in for better view.}
    \label{fig:ablation_study_for_merge_more}
\end{figure*}

\Fref{fig:ablation_study_for_merge_more} illustrates qualitative results showing how visual quality changes with different token compression ratios across sparse-view inputs. The figure indicates that for $1\sim4$ views a high merge ratio $r>0.5$ significantly degrades visual quality, whereas for $8\sim16$ views, larger merge ratios $r\in[0.5, 0.75]$ do not obviously reduce performance. We attribute this to the increased redundancy of tokens as the number of views grows, and the our model de facto does not require all tokens for effective reconstruction.

Moreover, \Tref{tab:ablation_study_for_merge_more} presents a quantitative ablation study of image-token merging. The table reports PSNR averaged across both public animation and smartphone-captured video datasets. the result shows that applying token merging with moderate ratios $r \in [0.25, 0.75]$ for sparse-view inputs $4\sim16$ yields metrics comparable to, or slightly better than, the no-merge baseline $r = 0$. These results are consistent with findings from previous work in image generation~\cite{bolya2023tomesd}.

\begin{table}[ht]
    \centering
    \caption{\textbf{Ablation study of image-token merging.} The PSNR metrics are averages computed over public and smartphone-captured video sequences.}
    \label{tab:ablation_study_for_merge_more} 

    \resizebox{0.48\textwidth}{!}{
    \large
    \begin{tabular}{*{7}{c}}
        \toprule
       Input Views & $r=0.00$ & $r=0.25$  &  $r=0.50$ & $r=0.75$  & $r=0.90$ & $r=0.95$ \\
       \midrule
       view = 1 & \textbf{27.762} & 27.735& 27.733 & 27.644 & 27.525 & 27.238 \\
       view = 4 & 27.942 & \textbf{27.942} & 27.940 & 27.937 & 27.893 & 27.662 \\
       view = 8 & 28.140 & \textbf{28.153} & 28.147 & 28.135 &27.903 & 27.735 \\
       view = 16 & 28.391 & \textbf{28.394} & \textbf{28.394} & 28.391 & 28.358 & 28.206\\
        \bottomrule
    \end{tabular}
    }
\end{table}

\subsection{Gaussian Regularization Loss} 
\label{sec:gaussian_regular}
\paragraph{Geometric Regularization} 
Video-based supervision frequently suffers from incomplete coverage of the human body and typically exhibits limited motion diversity. These limitations often lead to geometric artifacts and ambiguities in unobserved regions. To mitigate these issues, we incorporate two geometric regularizers adopted from LHM~\cite{qiu2025LHM}: (1) the \emph{As Spherical As Possible} loss, \(\mathcal{L}_{\text{ASAP}}\), which enforces isotropy in the 3D Gaussians; and (2) the \emph{As Close As Possible} loss, \(\mathcal{L}_{\text{ACAP}}\), which preserves spatial coherence among neighboring primitives.

\paragraph{Mask Distribution Regularization} 
Furthermore, empirical observations indicate that the optimization process tends to favor expanding Gaussian scales over refining positional offsets to represent geometry. This behavior induces significant artifacts, particularly when animating avatars with loose clothing in dynamic poses. To counteract this tendency, we propose a \emph{Mask Distribution Loss}, \(\mathcal{L}_{\text{dis}}\), which encourages Gaussian primitives to prioritize positional accuracy over scale expansion.

\begin{figure}[htb] 
    \centering
    \includegraphics[width=0.45\textwidth]{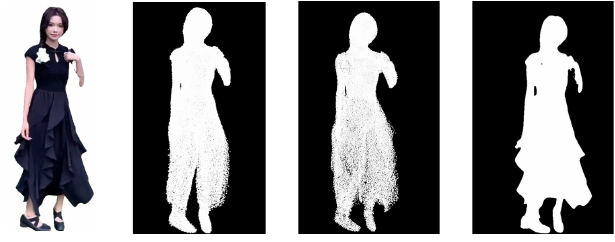}
    
    \makebox[0.11\textwidth][c]{\footnotesize Input} 
    \makebox[0.11\textwidth][c]{\footnotesize w/o \(\mathcal{L}_{\text{dis}}\)} 
    \makebox[0.11\textwidth][c]{\footnotesize w/ \(\mathcal{L}_{\text{dis}}\)} 
    \makebox[0.11\textwidth][c]{\footnotesize Human Mask} 
    \caption{Visual ablation study of the mask distribution loss. The comparison illustrates how \(\mathcal{L}_{\text{dis}}\) promotes a more uniform spatial distribution of Gaussian primitives, aligning better with the ground-truth human mask.} 
    \label{fig:ablation_for_mdl}
\end{figure}

Specifically, this regularization is implemented by rendering an auxiliary mask, \(M_{\text{dis}}\), using fixed Gaussian attributes (opacity \(\rho = 0.95\), scale \(\sigma = 0.002\)), and minimizing the L1 discrepancy between \(M_{\text{dis}}\) and the ground-truth human mask. By fixing the scale and opacity, this loss underscores the contribution of Gaussian positions. As demonstrated in \fref{fig:ablation_for_mdl}, incorporating this term encourages a more uniform distribution of Gaussian centers across the ground-truth mask area, effectively preventing the model from converging to solutions with excessively large scale parameters.

The combined geometric regularization term is formulated as:
\begin{equation}
    \mathcal{L}_{\text{reg}} = \lambda_{\text{dis}} \mathcal{L}_{\text{dis}} + \lambda_{\text{ASAP}} \mathcal{L}_{\text{ASAP}} + \lambda_{\text{ACAP}} \mathcal{L}_{\text{ACAP}},
\end{equation}
where the weighting coefficients are empirically set to \(\lambda_{\text{dis}} = 0.5\), \(\lambda_{\text{ASAP}} = 20\), and \(\lambda_{\text{ACAP}} = 5\).

\subsection{3D-Aware Human Animation Neural Renderer}
\label{sec:renderer}
To improve the rendering quality, we develop a lightweight neural renderer built on the 3DGS renderer to refine the final output. Specifically, we utilize the human prior model SMPL-X to animate the reconstructed 3D avatar to the target pose, then render high-dimension semantic features via 3DGS rasterization. Subsequently, we predict final render results using the lightweight neural renderer.

\paragraph{Rendering Time Across Different Resolutions}  The entire rendering process can be run in real time. \Tref{tab:experiments_for_render_time} clearly illustrates the rendering time of this process across different resolutions.

\begin{table}[htb]
    \centering
    \caption{Rendering Time of the 3D-aware neural render across different resolutions on a Single NVIDIA A100-80G}
    \label{tab:experiments_for_render_time}
    \resizebox{0.32\textwidth}{!}{
    \large
    \begin{tabular}{*{9}{c}}
        \toprule
        Resolution & Time (s) $\downarrow$ & FPS $\uparrow$ \\
        \midrule
        $592 \times 592$ & 0.013 & 78 \\
        $880 \times 880$ & 0.027 & 37 \\
        $1176 \times 1176$ & 0.048 & 21 \\
        \bottomrule
    \end{tabular}
    }
\end{table}

\paragraph{Effect of 3DGS Representation for Neural Renderer} 
Our renderer consists of a Gaussian Splatting rasterizer and a lightweight neural renderer. 
This raises an important question: Is the 3DGS intermediate representation necessary? Could the framework rely solely on a neural renderer, similar to HumanRAM~\cite{yu2025humanram}? 
To answer this question, we performed a comparison experiment to validate our hypothesis: \emph{the 3DGS representation plays a crucial role in the neural renderer}.

For the without-3DGS experiments, we employ point rasterization instead of the 3DGS rasterizer. 
Specifically, after the encoder-decoder PIT, we deform the points with high-dimensional semantic features and render point features via point rasterization implemented with PyTorch3D. 
Then, we use the neural renderer to obtain the final results. 
For training, apart from the 3DGS representation, all settings were identical to \MethodName.

\begin{figure*}[htb] 
    \centering
    \includegraphics[width=1.0\textwidth]{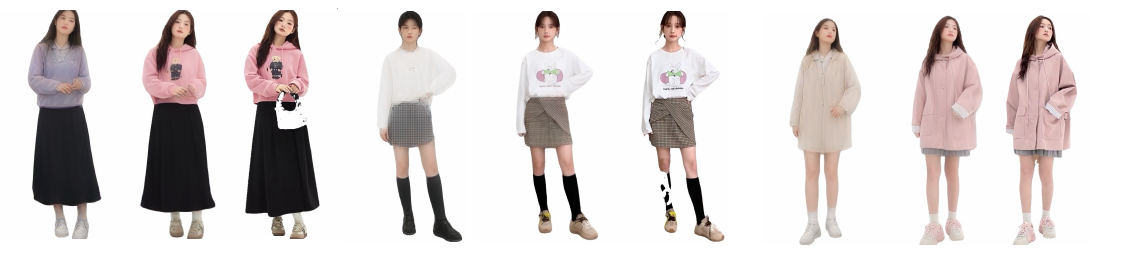} \\
    \makebox[0.1\textwidth][c]{\footnotesize w/o GS} 
    \makebox[0.1\textwidth][c]{\footnotesize w/ GS} 
    \makebox[0.1\textwidth][c]{\footnotesize GT} 
    \makebox[0.1\textwidth][c]{\footnotesize w/o GS} 
    \makebox[0.1\textwidth][c]{\footnotesize w/ GS} 
    \makebox[0.1\textwidth][c]{\footnotesize GT} 
    \makebox[0.1\textwidth][c]{\footnotesize w/o GS} 
    \makebox[0.1\textwidth][c]{\footnotesize w/ GS} 
    \makebox[0.1\textwidth][c]{\footnotesize GT} 
    \caption{The effects of 3DGS representation for neural renderer. `GS' indicates 3D Gaussian Splatting.} 
    \label{fig:ablation_for_dpthead}
\end{figure*}

As shown in \fref{fig:ablation_for_dpthead}, the comparison experiment illustrates that relying solely on point offsets without an explicit 3DGS representation causes the neural renderer to struggle with reproducing the input-image context, producing high-frequency artifacts and blurring. 
Moreover, the pipeline experiences convergence difficulties. 
In contrast, employing an explicit 3DGS representation enables faster convergence of the neural renderer and substantially improves overall visual quality.

\subsection{Implementation Details}
\label{sec:imple}

\paragraph{Training Strategy} 
To ensure training stability, we initially train \MethodName for single-view input by minimizing the training loss using the AdamW optimizer for 65~K iterations, excluding the 3D-aware neural renderer. 
Subsequently, to enable a pose-free setting capable of handling arbitrary image inputs, we randomly sample 1 to 16 frames from a training video for each batch. 
We then continue training for an additional 41~K iterations using the same optimizer and loss objectives. 
We apply gradient norm clipping with a threshold of 0.1 and employ bfloat16 precision with gradient checkpointing to optimize GPU memory usage and computational efficiency. 
A cosine learning rate scheduler is employed with a peak learning rate of $1\text{e-}4$ and a warm-up period of 3,000 iterations. 
Input images are resized to a maximum dimension of 1024 pixels. 
Training is conducted on 32 A100 GPUs for approximately one week.

\paragraph{Training Dataset} 
For training, we utilize approximately 300,000 in-the-wild video sequences collected from public repositories, complemented by over 5,173 synthetic static 3D human scans sourced from 2K2K~\cite{han2023highfidelity3dhumandigitization}, Human4DiT~\cite{shao2024human4dit}, and RenderPeople. 
Specifically, we sample training batches from the in-the-wild and synthetic datasets at a ratio of 19:1 to balance generalization and view consistency. 
To mitigate view bias in the video data, we sample perspectives uniformly across a diverse range, guided by the estimated global orientation of SMPL-X.

\section{Experiments}
\label{sec:experiments}

\subsection{Details of Public Animation Video} 
\label{sec:data}
We collect 23 public videos to form our evaluation benchmark. 
Specifically, we selected \emph{anran\_dance\_self\_rotated}, \emph{anran\_purple}, \emph{anran\_skirt}, \emph{lingteng}, \emph{self-rotate-leyang}, and \emph{xiaolin} from Rec-MV~\cite{qiu2023recmv}. 
Additionally, we sampled \emph{bike}, \emph{citron}, \emph{parkinglot}, \emph{jogging}, \emph{lab}, and \emph{seattle} from NeuMan~\cite{jiang2022neuman}. 
Furthermore, we included \emph{00000\_random}, \emph{00020\_Dance}, \emph{00069\_Dance}, \emph{exstrimalik}, \emph{helge}, \emph{lam}, \emph{Nadia}, \emph{phonecall}, \emph{roger}, \emph{truman}, and \emph{Yuliang} from Vid2Avatar~\cite{guo2023vid2avatar, guo2025vid2avatar}.


\begin{figure*}[htb] 
    \centering
    \includegraphics[width=0.96\textwidth]{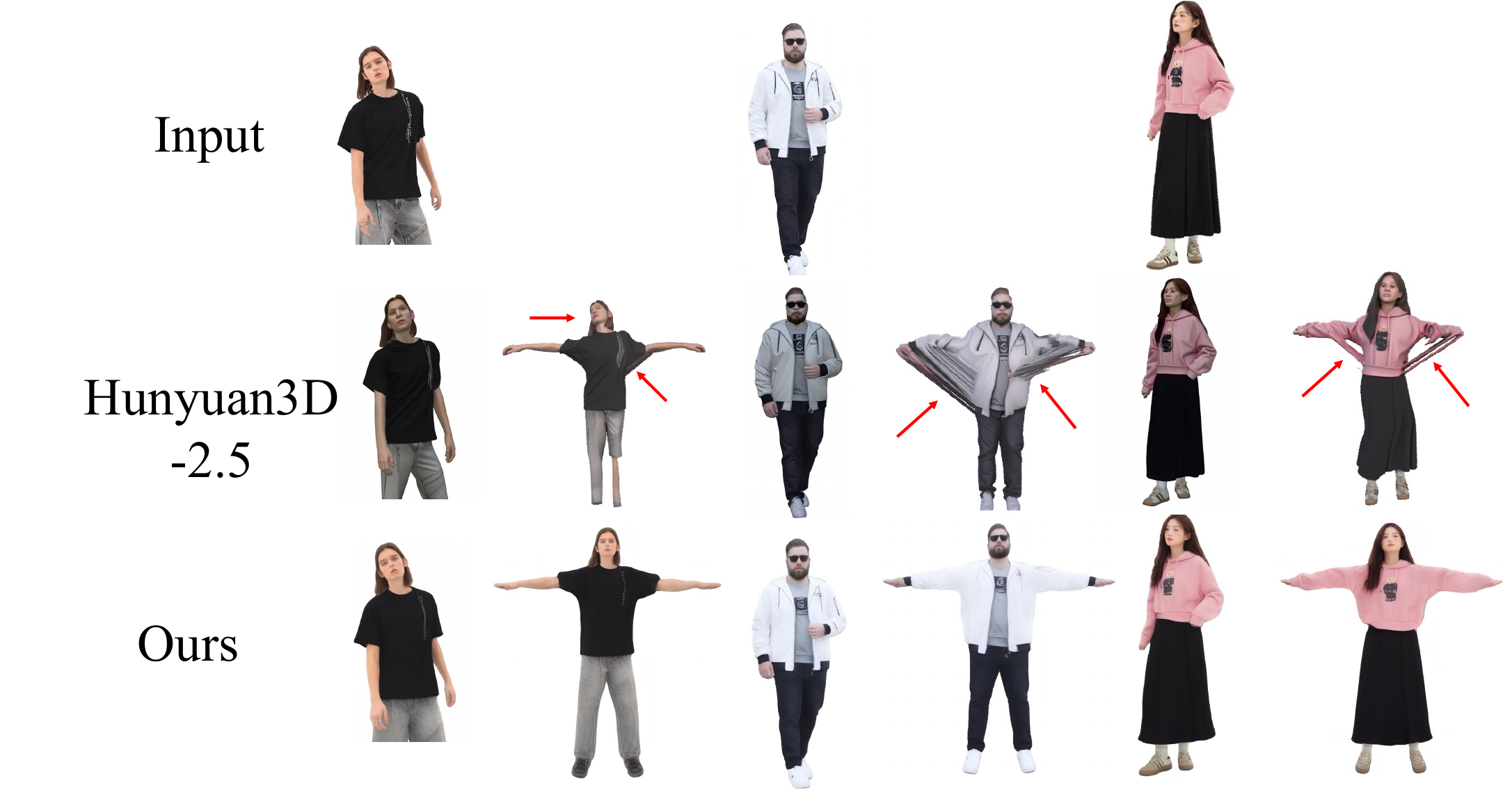}  \\
    \makebox[0.2 \textwidth][c]{\footnotesize} 
    \makebox[0.12\textwidth][c]{\footnotesize Recon} 
    \makebox[0.12\textwidth][c]{\footnotesize Reset} 
    \makebox[0.12\textwidth][c]{\footnotesize Recon} 
    \makebox[0.12\textwidth][c]{\footnotesize Reset} 
    \makebox[0.12\textwidth][c]{\footnotesize Recon} 
    \makebox[0.12\textwidth][c]{\footnotesize Reset} 
    \caption{\textbf{Comparison with Hunyuan3D-2.5}. We use Hunyuan3D-2.5 to reconstruct 3D models in the input-view space. We then auto-rig the models using Mixamo and reset the target-view pose into the canonical space. "Recon" denotes reconstruction in the input space, while "Reset" denotes animating the avatar into a predefined canonical pose.}
    \label{fig:ablation-for-image-to-3D}
\end{figure*}

\subsection{Comparison with General Image-to-3D Generation}
\label{hunyuan}
We conducted comparative experiments against state-of-the-art general image-to-3D generation methods. 
These include cutting-edge commercial software, such as Rodin~\cite{zhang2024clay} and Hunyuan3D-2.5~\cite{lai2025hunyuan3d25}. 
For our experiments, we selected Hunyuan3D-2.5 as the baseline due to its state-of-the-art performance. 
Since Hunyuan3D only reconstructs static 3D assets, we use Mixamo to auto-rig the reconstructed models after 3D avatar reconstruction, making reconstructed static 3D avatar animatable.

As shown in \fref{fig:ablation-for-image-to-3D}, successful rigging of a 3D digital avatar demands stringent input pose requirements. 
Although Hunyuan3D-2.5 can recover the overall structure from the input image, it often fails to produce an ideal riggable geometry. 
This is evident in the "Reset" panel, where anatomical points on the arms and torso are difficult to distinguish, complicating avatar animation. 
Furthermore, the generated textures often appear unnatural, containing artifacts and unrealistic details. 
In contrast, LHM++ generates natural, high-fidelity avatars in canonical space with superior rigging compatibility.

\subsection{Scalability Analysis}
\label{sec:scalability}

\begin{figure}[htb] 
    \centering
    \includegraphics[width=0.45\textwidth]{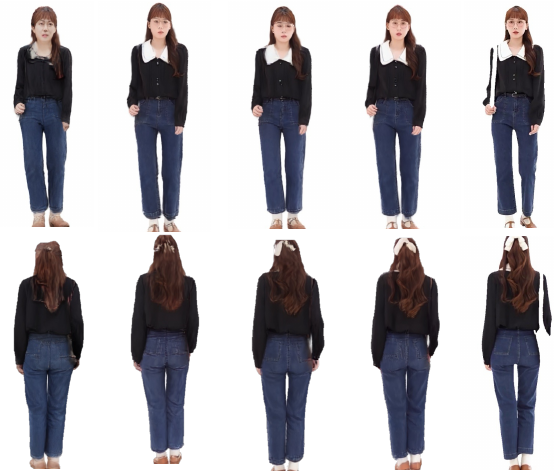}  \\
    \makebox[0.08\textwidth][c]{\footnotesize Synthetic} 
    \makebox[0.08\textwidth][c]{\footnotesize 10K} 
    \makebox[0.08\textwidth][c]{\footnotesize 100K} 
    \makebox[0.08\textwidth][c]{\footnotesize 300K} 
    \makebox[0.08\textwidth][c]{\footnotesize GT} 
    \caption{Ablation study on dataset scalability.} 
    \label{fig:dataset_scalability}
\end{figure}

\paragraph{Dataset Scalability} 
To assess dataset scalability, we performed controlled experiments using stratified random subsets of 10K and 100K videos from the original 300K training dataset. 
\Tref{tab:ablation_for_data} demonstrates that relying solely on synthetic data leads to poor model generalization. 
In contrast, incorporating in-the-wild data significantly improves model robustness and performance on real-world evaluations. 
Furthermore, increasing the dataset size yields consistent performance gains. 
\Fref{fig:dataset_scalability} illustrates the visual quality improvements as the number of in-the-wild videos increases.

\paragraph{Model Parameter Scalability} To verify the scalability of our \MethodName, we train variant models with increasing parameter numbers by scaling the layer numbers. Our experiments indicate that increasing the number of model parameters correlates with improved performance. However, the performance improvements tend to saturate as the model size continues to grow. We believe that this limitation is largely influenced by the scale of our training dataset that is extremely smaller—by an order of magnitude—than the extensive datasets employed in LLMs. Our experiments reveal that while performance shows moderate enhancements from small to medium-sized models, the gains become negligible for larger models.

\begin{table}[htb]
    \centering
    \caption{\textbf{Ablation study on dataset scalability.} The metrics are calculated based on 16-view image inputs on the in-the-wild video dataset.}
    \label{tab:ablation_for_data}
    \resizebox{0.48\textwidth}{!}{
    \large
    \begin{tabular}{*{4}{c}}
        \toprule
        Methods & PSNR $\uparrow$ & SSIM $\uparrow$ & LPIPS $\downarrow$ \\
        \midrule
        Synthetic Data & 20.549 & 0.858 & 0.086 \\
        10K Videos & 21.526 & 0.870 & 0.073 \\
        100K Videos & 22.051 & 0.885 & 0.062 \\
        300K Videos + Synthetic Data & \textbf{22.354} & \textbf{0.887} & \textbf{0.057} \\
        \bottomrule
    \end{tabular}
    }
\end{table}

\subsection{Training and Inference Efficiency} 
\label{sec:efficiency}
\Tref{tab:model_efficiency} reports end-to-end training and inference times, including preprocessing and image tokenization. 
The results show that the proposed method significantly reduces both training/inference time and memory consumption. 
During training, LHM encounters out-of-memory (OOM) errors when handling 160K query points and requires approximately 16$\times$ more time per iteration compared to our framework. 
Regarding inference, our method delivers a 55$\times$ speedup compared to LHM with the same number of query points while consuming only 6--8GB GPU memory, making \MethodName deployable on consumer devices.

\begin{table}[tb]
    \centering
    \caption{\textbf{End-to-end efficiency.} Batch size=2, input views=8. \# Points = geometric points; Time = per-iteration duration (training/inference).}
    \label{tab:model_efficiency}



\resizebox{0.48\textwidth}{!}{
\large
\begin{tabular}{l c  c | cc | cc}
\toprule
\multirow{2}{*}{\textbf{Methods}} & 
\multirow{2}{*}{\textbf{Params.}} & 
\multirow{2}{*}{\textbf{\# Points}} & 
\multicolumn{2}{c}{\textbf{Training}} & 
\multicolumn{2}{|c}{\textbf{Inference}} \\
& & & Time & Memory & Time & Memory  \\
\midrule
\multirow{2}{*}{LHM-0.7B*~\cite{qiu2025LHM}} & 700~M & 40~K & 81.22~s & 80.3~GB & 24.76~s& 25.42~GB \\
& 700~M & 160~K & - & OOM & 110.3~s & 36.71~GB \\
\midrule
\MethodName~Small & 500~M & 160~K & 5.79~s & 43.5~GB & 1.58~s & 6.98~GB \\
\MethodName~Medium & 700~M & 160~K & 6.35~s & 48.7~GB & 2.09~s & 7.35~GB \\
\MethodName~Large & 900~M & 160~K & 7.25~s & 52.6~GB & 2.88~s & 8.21~GB\\
\bottomrule
\end{tabular}
}

\end{table}

\subsection{More Quantitative Results}
\label{sec:quan}
Table~3  of the main paper summarizes the averaged results on the public and smartphone-captured video sequences. \Tref{tab:animation_public} and \Tref{tab:animation_casual} show the detailed results for each dataset.

\begin{table*}[htb]\centering
    \caption{Comparison experiments with sparse-view input methods on public animation dataset.
    }
    \label{tab:animation_public}

\resizebox{0.96\textwidth}{!}{
\large
\begin{tabular}{c| cccc | cccc | cccc | cccc}
\toprule
\multirow{2}{*}{\textbf{views}} & 
\multicolumn{4}{c|}{\textbf{InstantAvatar}~\cite{jiang2023instantavatar}} & 
\multicolumn{4}{c|}{\textbf{GaussianAvatar}~\cite{hu2024gaussianavatar}} & 
\multicolumn{4}{c|}{\textbf{ExAvatar}~\cite{moon2024exavatar}} & 
\multicolumn{4}{c}{\textbf{LHM++} M} \\
& PSNR & SSIM & LPIPS & Time & PSNR & SSIM & LPIPS & Time & PSNR & SSIM & LPIPS & Time & PSNR & SSIM & LPIPS & Time \\
\midrule
2  & 21.718 & 0.929 & 0.125 &  3.6~m & 21.049 & 0.937 & 0.086  & 3.8~m &  26.371 & 0.959 & 0.045 & 8.5~m & \textbf{26.900} & \textbf{0.964} & \textbf{0.022} & \textbf{1.08~s} \\
4  & 21.782 & 0.926 & 0.126 & 6.0~m & 22.614 & 0.943 & 0.065 &  5.6~m &  26.906 & 0.960 & 0.044 & 15~m & \textbf{27.031} & \textbf{0.965} & \textbf{0.022} & \textbf{1.40~s} \\
8  & 22.308 & 0.928 & 0.120 & 10.6~m & 23.238 & 0.947 & 0.061 &  9.0~m & 27.319 & 0.962 & 0.040 & 32~m &  \textbf{27.346} &\textbf{ 0.965} & \textbf{0.021} & \textbf{2.09~s} \\
16 & 23.026 & 0.936 & 0.101 & 18.8~m & 23.427 & 0.949 & 0.061 &  15.7~m & 27.456 & 0.962 & 0.040 & 1.2~h &  \textbf{27.521} & \textbf{0.966} & \textbf{0.020} & \textbf{3.66~s} \\
64 &  23.561 & 0.944 & 0.077 & 54.0~m & 23.574 & 0.949 & 0.055 & 1.1~h & \textbf{27.896} & 0.962 & 0.037 & 4.4~h & 27.557 & \textbf{0.966} & \textbf{0.020} & \textbf{17.94~s} \\
\bottomrule
\end{tabular}
}
    \vspace{-0.5em}
\end{table*}

\begin{table*}[htb]\centering
    \caption{Comparison experiments with sparse-view input methods on smartphone-captured videos.
    }
    \label{tab:animation_casual}
    \resizebox{0.96\textwidth}{!}{
\large
\begin{tabular}{c| cccc | cccc | cccc | cccc}
\toprule
\multirow{2}{*}{\textbf{views}} & 
\multicolumn{4}{c|}{\textbf{InstantAvatar}~\cite{jiang2023instantavatar}} & 
\multicolumn{4}{c|}{\textbf{GaussianAvatar}~\cite{hu2024gaussianavatar}} & 
\multicolumn{4}{c|}{\textbf{ExAvatar}~\cite{moon2024exavatar}} & 
\multicolumn{4}{c}{\textbf{LHM++ M}} \\
& PSNR & SSIM & LPIPS & Time & PSNR & SSIM & LPIPS & Time & PSNR & SSIM & LPIPS & Time & PSNR & SSIM & LPIPS & Time \\
\midrule
2  & 24.340  & 0.954 & 0.070 &  {3.6~m} & 22.602 & 0.953 & 0.056 & 3.8~m &  {26.723} &  {0.969} &  {0.031} & 8.5~m &  \textbf{28.413} &  \textbf{0.974} & \textbf{0.016} &  \textbf{1.08 s} \\
4  & 24.387 & 0.953 & 0.072 & 6.0~m & 23.857 & 0.957 & 0.043 & 5.6~m &  {27.532} &  {0.970} &  {0.028} & 15~m &  \textbf{28.502} & \textbf{0.974} & \textbf{0.016} &  \textbf{1.40~s} \\
8  & 24.481 & 0.952 & 0.072 & 10.6~m & 24.644 & 0.961 & 0.040 &  {9.0~m} &  28.431 &  {0.972} &  {0.026} & 32~m &  \textbf{28.643} &  \textbf{0.974} &  \textbf{0.015} &  \textbf{2.09~s} \\
16 & 24.984 & 0.957 & 0.059 & 18.8~m & 24.999 & 0.962 & 0.039 &  {15.7~m} &  28.916 &   {0.973} &  {0.024} & 1.2~h &  \textbf{28.934} &  \textbf{0.975} &  \textbf{0.014} &  \textbf{3.66~s} \\
64 &  26.329 & 0.964 & 0.042 & 54.0~m & 26.381 & 0.967 & 0.030 & 1.1~h & \textbf{30.634} & \textbf{0.978} & 0.020 & 4.4~h & 28.958 & 0.975 & \textbf{0.013} & \textbf{17.94~s}\\
\bottomrule
\end{tabular}
}

\end{table*}

\subsection{More Qualitative Results}
\label{sec:qual}

\begin{figure*}[tb] \centering
    \includegraphics[width=0.96\textwidth]{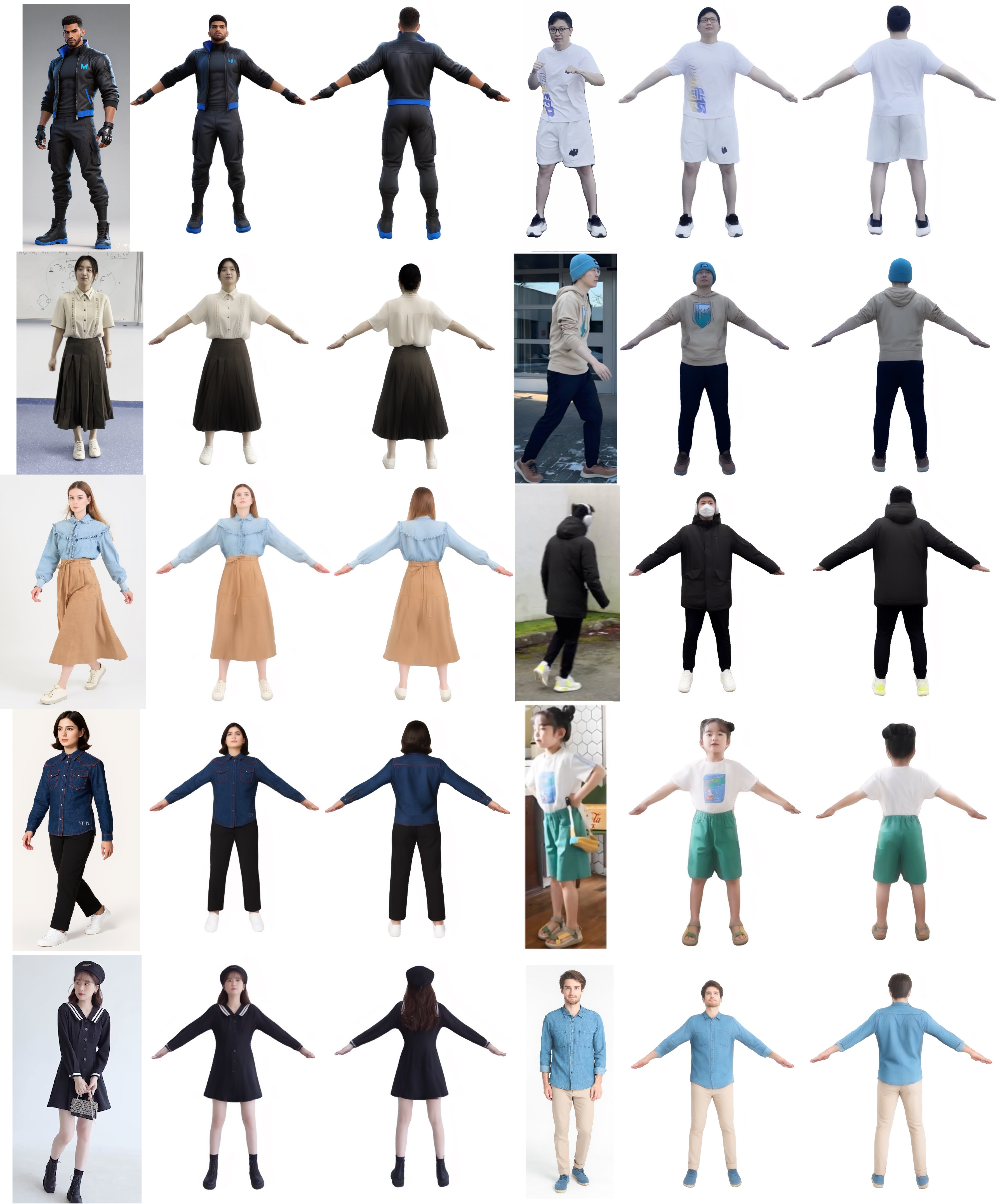}
    \\
    \makebox[0.1\textwidth]{\footnotesize Reference}
    \makebox[0.35\textwidth]{\footnotesize Reconstruction}
    \makebox[0.1\textwidth]{\footnotesize Reference}
    \makebox[0.35\textwidth]{\footnotesize Reconstruction}
    \caption{Human reconstruction in canonical space from 8 input images. The reference image is one of the inputs.}
    \label{fig:canonical_avatar}
\end{figure*}

\paragraph{Human Reconstruction in Canonical Space} \Fref{fig:canonical_avatar} demonstrates human reconstructions in canonical space from sparse views, showing that our model can reconstruct reasonable, high-quality avatars from sparse images input.

\begin{figure*}[tb] \centering
    \includegraphics[width=0.9\textwidth]{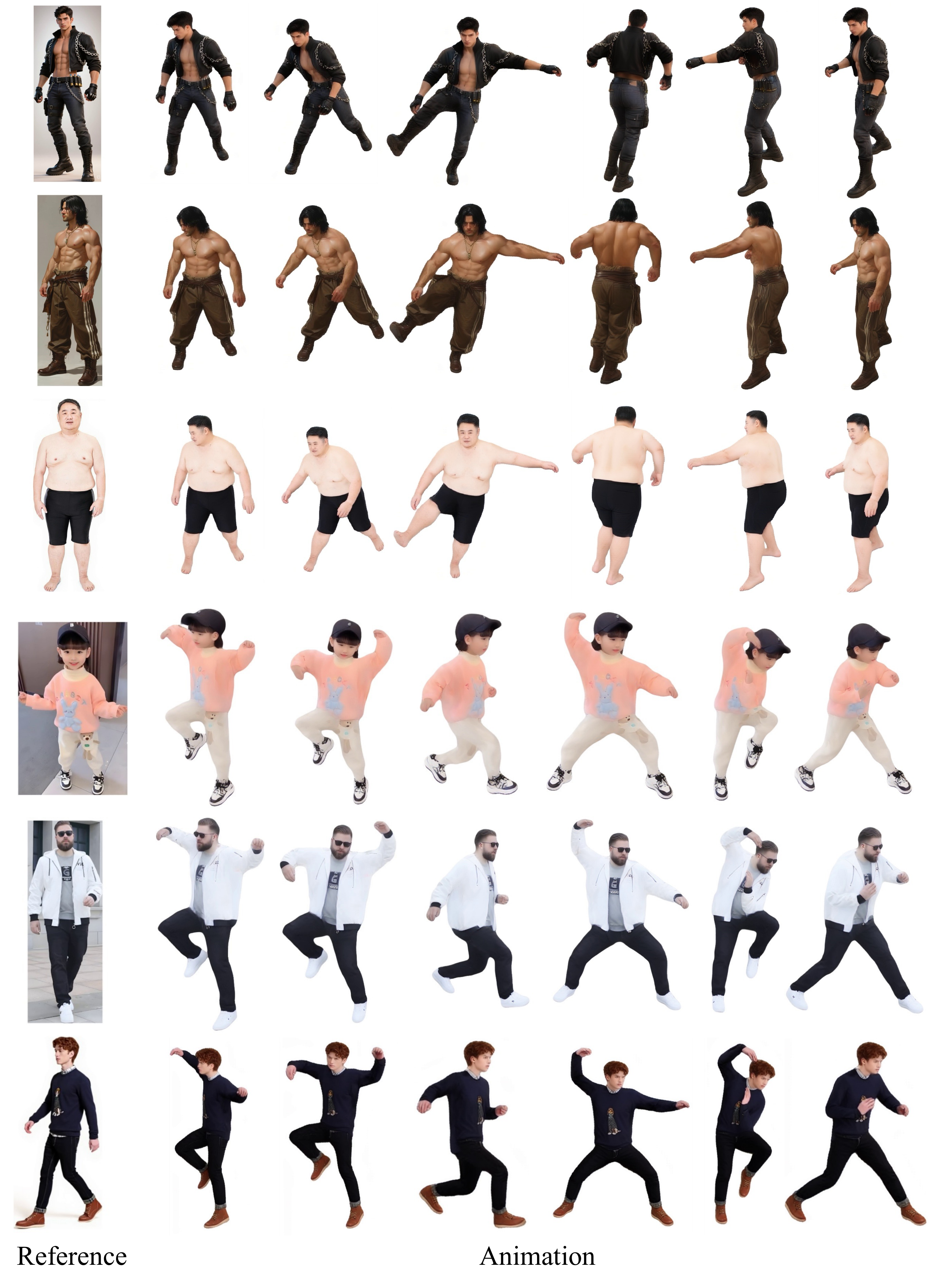}
    \\
    \vspace{-0.5em}
    \caption{More animation results of avatars created with 8-image inputs~(Part I). \textbf{Reference} image is one of the input images.} \label{fig:supp_animation_1}
\end{figure*}

\begin{figure*}[tb] \centering
    \includegraphics[width=0.8\textwidth]{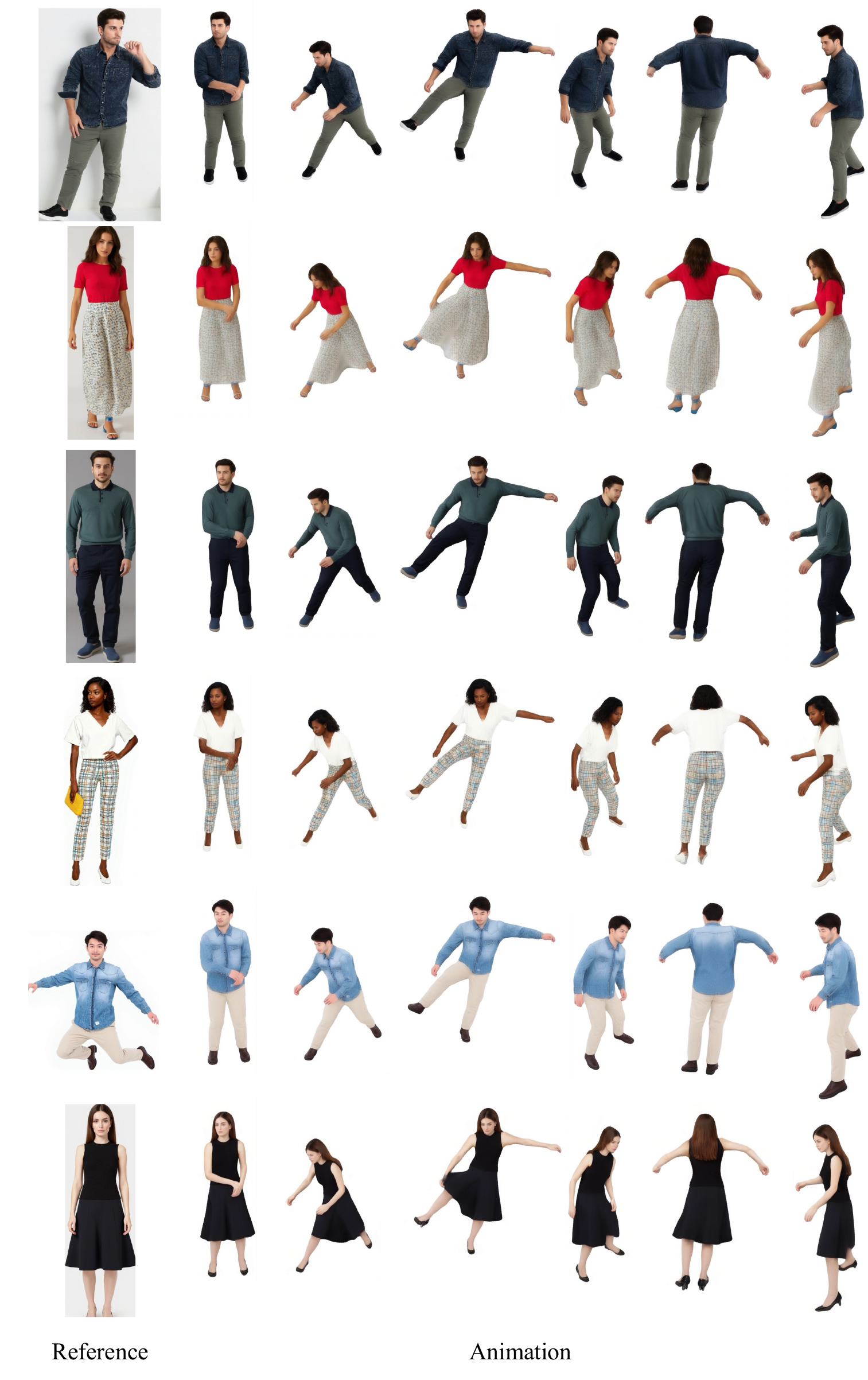}
    \\
    \vspace{-0.5em}
    \caption{More animation results of avatars created with 8-image inputs~(Part II). \textbf{Reference} image is one of the input images.} \label{fig:supp_animation_2}
\end{figure*}

\paragraph{More Animation Rendering Results} Both \fref{fig:supp_animation_1} and \fref{fig:supp_animation_2}  present more animation results using eight input images. Our method enables high-fidelity reconstruction and animation of human avatars through the efficient Point-Image Transformer architecture, thereby demonstrating robust generalization capabilities and practical effectiveness.


\section{Limitations and Future Work} 
\label{sec:limitation}

\begin{figure*}[htb] \centering
    \includegraphics[width=0.96\textwidth]{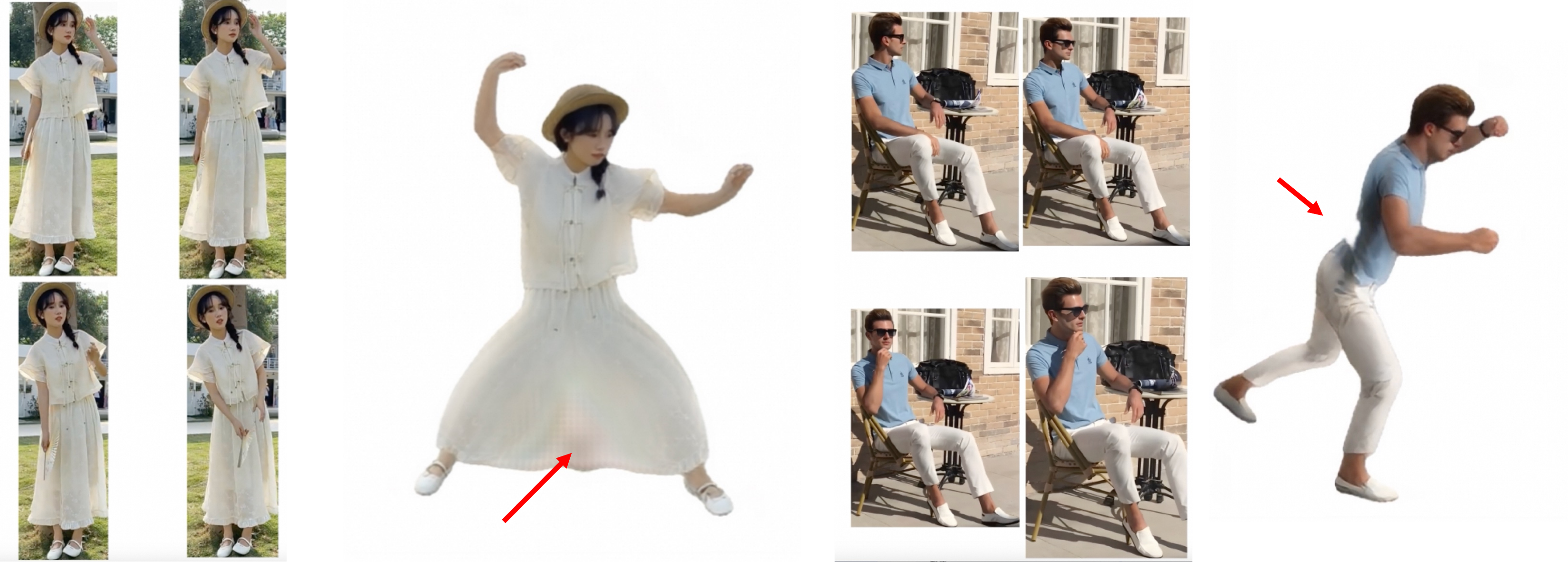} \\
    \makebox[0.24\textwidth][c]{\footnotesize Inputs} 
    \makebox[0.24\textwidth][c]{\footnotesize Animation Results} 
    \makebox[0.24\textwidth][c]{\footnotesize Inputs} 
    \makebox[0.24\textwidth][c]{\footnotesize Animation Results} 
    \caption{\textbf{Failure Cases.} our model struggles to animate avatars wearing skirts during large-amplitude lower‑body movements, and has difficulty handling inputs with extreme or challenging poses.}
    \label{fig:failure-case}
\end{figure*}

A primary limitation of our approach is its dependence on the SMPL-X template mesh to initialize geometric tokens. While this provides a strong structural prior, it can limit reconstruction fidelity for subjects wearing loose or non-body-conforming garments, such as dresses and skirts. As shown in \Fref{fig:failure-case}, this manifests in failure cases where the model struggles to animate avatars wearing skirts during large lower-body movements. Furthermore, due to the limited diversity of large-motion poses in our training datasets, the model's performance may decline when faced with unseen or extreme poses, which is also reflected in the failure cases presented. Additionally, since our model is primarily designed for sparse view input, we find that as the number of input views exceeds 16, the improvement in performance becomes marginal. 

In future work, we plan to investigate more flexible garment representations and explore pose-independent anchor structures to better capture complex clothing dynamics, enhancing generalization to diverse motions. Moreover, our aim is to design a framework that is more suitable for dense view inputs.

\end{document}